\pdfoutput=1

\documentclass[11pt]{article}

\usepackage{ifthen} 
\newboolean{review}
\setboolean{review}{false}  
\newboolean{CR}
\setboolean{CR}{true}

\newcommand{\iffinal}[2]{%
  \ifthenelse{\not\boolean{review}}{#1}{#2}%
}

\newcommand{\repolink}
    {\href{https://github.com/jsrozner/perturbation}
    {github.com/jsrozner/perturbation}.
}

\iffinal
    {\ifthenelse{\boolean{CR}}
        {\usepackage[final]{acl}}
        {\usepackage[preprint]{acl}}
    }
    {\usepackage[review]{acl}}

\usepackage{times}
\usepackage{latexsym}
\usepackage[T1]{fontenc}
\usepackage[utf8]{inputenc}
\usepackage{microtype}
\usepackage{inconsolata}
\usepackage{graphicx}
\usepackage{listings}
\usepackage{overpic}
\usepackage{makecell}

\usepackage{amsmath}
\usepackage{amsfonts}
\usepackage{xcolor}

\usepackage{hyperref} 
\usepackage{xstring}  
\usepackage{tabularx}
\usepackage{pgf}
\usepackage{subcaption}
\usepackage{graphicx}
\usepackage{adjustbox}

\usepackage{authblk}

\newcommand{\ttw}[1]{\textit{#1}}

\newcommand{\sectionref}[1]{%
  \texorpdfstring{%
    \hyperref[#1]{Section~\ref*{#1}: \nameref*{#1}}%
  }{%
    Section~\ref*{#1}%
  }%
}
\newcommand{\sectionnameref}[1]{\hyperref[#1]{\nameref{#1}}}
\newcommand{\sref}[1]{\S\ref{#1}}
\newcommand{\aref}[1]{\ref{#1}}
\newcommand{\figref}[2]{\hyperref[#1]{Fig \ref{#1}#2}}

\usepackage{multirow} 
\usepackage{nicematrix} 
\usepackage{booktabs}

\usepackage{relsize}
\usepackage{floatrow}
\DeclareFloatFont{tiny}{\small}
\usepackage{tcolorbox}       
\usepackage{multicol}        
\usepackage{afterpage}

\usepackage{enumitem}
\setlist{noitemsep}

\usepackage{comment}

\usepackage{xcolor}
\usepackage{soul}

\usepackage{xspace}

\newcommand{\perturb}[4]{\ul{\textit{#1 \textbf{#2} #3} [#2 $\rightarrow$ #4]}}
\newcommand{\minp}[6]{\textit{#1 \textbf{[#2, #3]} #4 $\rightarrow$ \textbf{[#5, #6]}}}

\newcommand{\crs}{\emph{critical region similarity}\xspace}
\newcommand{\crshort}{CRS\xspace}
\newcommand{\ctla}{\emph{ctrl-agr}\xspace}
\newcommand{\ctlt}{\emph{ctrl-trans}\xspace}
\newcommand{\margeff}{\texttt{marginal-effects}\xspace}

\newcommand{\blt}{\emph{baselined-transfer}\xspace}
\newcommand{\Blt}{\emph{Baselined-transfer}\xspace}

\renewcommand*{\Affilfont}{\normalsize\normalfont}

\title{Perturbation: A simple and efficient adversarial tracer \\ for representation learning in language models}

\makeatletter
\renewcommand\AB@affilsepx{\hspace{1em} \protect\Affilfont}
\makeatother

\author[1]{Joshua Rozner}
\author[1]{Cory Shain}

\affil[1]{Stanford University\protect\\
\texttt{
rozner@stanford.edu,
cory.shain@gmail.com
}}

\begin{document}
\maketitle

\begin{abstract}
Linguistic representation learning in deep neural language models (LMs) has been studied for decades, but
finding representations in LMs remains an unsolved problem.
On the one hand, unconstrained alignments may trivialize the notion of representation \cite{sutter2025non};
on the other, even recently popularized \emph{linear} approaches may not always be faithful to natural model behavior
\cite{arora-etal-2024-causalgym}.
Here we escape this dilemma by reconceptualizing representations not as patterns of activation but as conduits for learning.
Our approach is simple: we perturb an LM by fine-tuning it on a single adversarial example and measure how this perturbation ``infects'' other examples.
Perturbation makes no geometric assumptions, and unlike other methods, it does not find representations where it should not (e.g., in untrained LMs). 
But in trained LMs, perturbation reveals structured transfer at multiple linguistic grain sizes, suggesting that LMs both generalize along representational lines and acquire linguistic abstractions from experience alone.
\end{abstract}

\section{Introduction}
\label{sec:intro}

Understanding the latent representations of black-box machine learning systems like neural language models (LMs) is a growing scientific field \cite{smolensky_tensor_1990,mccoy2018rnns,pimentel-etal-2020-information,rogers-etal-2020-primer,belinkov-2022-probing,geiger2025causal}, and many studies in this domain have argued that LMs acquire a rich diversity of linguistic abstractions during training, ranging from morphology \cite{mikolov2013b,stanczak-etal-2022-neurons} to syntax \cite{linzen-etal-2016-assessing,tenney-etal-2019-bert,hewitt-manning-2019-structural,gauthier-etal-2020-syntaxgym,hu2024} to semantics \cite{conneau-etal-2018-cram,Templeton2024-hs}.
For this reason, LMs are a powerful testbed for longstanding scientific questions about the kinds of linguistic representations that can be acquired from experience \cite{gold1967language, elman1990finding, warstadt2022artificial, misra-mahowald-2024-language, 
milliere2024LMs, rozner-etal-2025-constructions, futrell_mahowald_2025_linguists}.

Finding representations in LMs remains an open problem.
Early work hypothesized that representations would be encoded as activation patterns and attempted to find representations in specific neurons \cite{linzen-etal-2016-assessing} or by training probing classifiers \cite{tenney-etal-2019-bert}.
Such probing approaches are fraught \cite{alain2017understanding}: they are correlational \cite{geiger2022inducing}, 
they can learn representations directly \cite{hewitt-liang-2019-designing, belinkov-2022-probing}, 
and they will fail to find representations outside the classifier's solution space \cite{adi2017finegrained}.
There has thus been a shift toward approaches, collectively called \textit{mechanistic interpretability} \cite{Elhage2021-qh}, that attempt to ground the notion of representation in the causal effects of activation patterns on model behavior \cite{tucker-etal-2021-modified,Templeton2024-hs,saphra-wiegreffe-2024-mechanistic,geiger2025causal}.

Mechanistic interpretability generally has two goals: 
scientific inquiry---testing hypotheses about the representations underlying model behavior---and model control---finding ways to steer model behavior \cite{mueller-etal-2026-quest}. 
We study the former: the use of idealized statistical learners (LMs) to test psycholinguistic theories of language representations
\cite{jackendoff2017defense, branigan-pickering-2017}.
For this goal, an important attribute of methods is \emph{faithfulness}:  
distinguishing representations used by the model from structure induced by the analysis itself \cite{jacovi-goldberg-2020-towards}.

\paragraph{Limitations of supervised interpretabiliy methods for scientific inquiry}
One such method is distributed alignment search (DAS; \citealt{geiger2024finding}), 
which uses supervision to find a one-dimensional (linear) subspace 
that achieves a hypothesized causal abstraction \cite{wu2023interpretability, arora-etal-2024-causalgym}.
DAS is an increasingly popular method for characterizing linguistic representations in LMs 
\cite[e.g.,][]{boguraev-etal-2025-causal, boguraev2026drawbridge, desai-nair-2026-filling, steele2026das, trinley2026das},
though \citet{kumon-yanaka-2026-fine} recently suggest that it can overfit on narrow aspects of linguistic behavior.

DAS and related methods often draw on the \textit{linear representation hypothesis} that many model representations are approximately linearly encoded \cite{mikolov2013b,Elhage2022-is,park2024linear}.
But even linear DAS recovers spurious structure in untrained models 
(\citealt{wu2023interpretability, arora-etal-2024-causalgym}; 
see \textbf{\sectionnameref{sec:bg}}).
Additionally, although some LM representations have been shown to be linearly encoded \cite{nanda-etal-2023-emergent, park2024linear},
a number of known exceptions \cite[e.g.,][]{white-etal-2021-non, csordas-etal-2024-recurrent, diego2024polar, engels2025not} motivate relaxing the linearity constraint to accomodate other geometries.
However, relaxation of constraints on an alignment in turns risks recovering vacuous mechanisms \cite{sutter2025non}.

This presents a dilemma: linear DAS is both too expressive (capable of finding causal linear subspaces where it should not---e.g., in untrained models) and not expressive enough (incapable of finding nonlinear structure where it should).
Any attempt to fix one horn of the dilemma worsens the other.
Moreover, that DAS recovers causally effective mechanisms in untrained models suggests that techniques like DAS, which directly optimize for a causal effect, run the risk of reifying assumptions. 

\paragraph{Grounding causal representations in learning}
Here we propose to escape this dilemma via the insight that representations are conduits for learning \cite{hinton1986learning,shepard1987toward,Tenenbaum2011-ev}: 
changes to a representation generalize to all future uses of that representation.
We build on this idea with the method of \emph{perturbation}: we finetune a language model on a single adversarial example and trace how this corruption propagates to other examples.
Perturbation makes no assumptions about the representational geometry of the model's hidden states, but instead rests on the following key assumption: 
if a model represents a certain abstraction, then a corruption to that abstraction should selectively propagate to other examples that also implicate the abstraction.

We bring our method to bear on the aforementioned learnability questions in linguistic theory by applying it to the well-studied problem of whether and how LMs come to represent linguistic abstractions like syntactic categories and constructions \cite{kim-smolensky-2021-testing, misra-mahowald-2024-language, patil-etal-2024-filtered}.
There are many existing linguistic benchmarks that could be tested using our approach.
To keep things feasible, we selected three benchmarks of different linguistic grain sizes: morphological, lexical, and syntactic (see \sref{sec:methods:exp}).
Across these evaluations, we find that perturbation is both a sensitive and selective tool for causal representation discovery, finding rich interpretable linguistic structure in trained models and no structure in untrained ones.
This result supports both the use of perturbation as an LM interpretability technique and the hypothesis that LMs emergently discover rich linguistic abstractions from experience alone, convergent with many similar recent findings \cite[e.g.,][]{warstadt-etal-2023-findings,misra2023abstraction,hu2024,rozner-etal-2025-babylms}. 

\paragraph{Contributions}
We introduce \emph{perturbation}, an efficient, geometry-free interpretability method for finding linguistic representations.
To our knowledge, perturbation is the first method to ground the notion of representation in its (causal) effect on learning, and perturbation is thus unique in its ability to identify representations from a single example. 
As it is \emph{unsupervised},
perturbation addresses many of the \emph{faithfulness} concerns with that have been raised about supervised methods like DAS.

\section{Background}
\label{sec:bg}

\paragraph{Criticisms of DAS}
As discussed in the \textbf{Introduction}, DAS has become a popular method for finding linguistic abstractions.
However, because representation-finding techniques like DAS and probing train against the same metric against which they also evaluate, they risk recovering the researcher’s hypothesis rather than the representations of interest \cite{hewitt-liang-2019-designing}.
The faithfulness of supervised methods like DAS has been challenged in numerous recent papers 
(\citealt{makelov2024illusion, meloux2025everything, sutter2025non, grant2025divergent, kumon-yanaka-2026-fine}; cf. \citealt{wu2024replymakelove}).

Additionally, in their original work that established DAS as a tool for the study of linguistic representations, \citet{arora-etal-2024-causalgym} noted that ``\citet{wu2023interpretability} found [D]AS achieves substantial causal effect even on randomly-initialized models or with irrelevant next-token labels, both settings where no causal mechanism should exist,'' and thus added control tasks (see CausalGym §5.2). 
Importantly, the CausalGym appendix shows that, after controlling for excess expressivity, DAS---the only method that directly supervises for a causal effect---is on average \emph{less} selective than other methods that do not optimize for the effect under study, including even than the far more interpretable difference in means \cite{marks2024means}.
These concerns motivate the development of \emph{unsupervised} representation-finding methods.

\section{Methods}
\label{sec:methods}

\subsection{Perturbation}

We develop the method of \emph{perturbation} to quantify representational similarities across strings that contain a hypothesized representation.
For example, the word \ttw{duck} has multiple \emph{senses} (e.g., an animal, a food, and an action) that might be represented differently in models.
We therefore train on a single datapoint 
designed to corrupt a representation of interest (e.g., finetuning to generate \textit{glam} instead of \textit{duck} in the context \textit{A \_\_\_ quacks}) and then measure the impact of this intervention on other model behaviors.
For example, the context in this example favors the animal sense of \textit{duck} (as opposed to the food or action senses).
If the model is perturbed to produce \textit{glam} in such a context, in which other contexts will this perturbation increase the probability of \textit{glam} relative to \textit{duck}?
For all instances of \textit{duck} regardless of sense, for only the animal sense of \textit{duck}, or in another manner?
Perturbation thus becomes a kind of ``tracer injection'' for latent LM representations (e.g., putative word senses), allowing us to use gradations of transfer learning as a window into the underlying representation space.
Below we formalize this method by first defining the construct of a \textit{remapping} and then elaborating on how this construct is used both to define a perturbation and to evaluate its effects.

\paragraph{Remapping}
Let a \textit{token} be a pair $\langle w, i \rangle$ representing an element $w$ of the model's vocabulary $W$ at position $i \in \mathbb{N}$. 
Let a \textit{string} $S$ be a finite set of tokens $\{ \langle w^{(1)}, i^{(1)} \rangle, \ldots, \langle w^{(n)}, i^{(n)} \rangle \}$ with all position indices distinct, i.e.\ $i^{(j)} \neq i^{(k)}$ for $j \neq k$. 
Any string $S$ can be partitioned into a \textit{critical region} $R$ and a \textit{context} $C$.

Let a \emph{remapping} from an \emph{original} string $S_o = R_o \cup C_o$ to an \emph{alternate} string $S_a = R_a \cup C_a$ be a 4-tuple: $\langle C_o, R_o, C_a, R_a \rangle$.
In other words, a remapping expresses a correspondence between the contexts and critical regions of two strings (the original and the alternate).
An example remapping is $\langle \{\langle \text{A}, 1 \rangle, \langle \text{quacks}, 3 \rangle\},\allowbreak \{ \langle \text{duck}, 2 \rangle \},\allowbreak \{\langle \text{A}, 1 \rangle, \langle \text{quacks}, 3 \rangle\},\allowbreak \{ \langle \text{glam}, 2 \rangle \} \rangle$.
This example remaps \textit{A \textbf{duck} quacks} into \textit{A \textbf{glam} quacks}.
For simplicity, we will notate the remapping for the above example as follows, with the critical region in \textbf{bold}: 
\perturb{A}{duck}{quacks}{glam}.
We use the construct of a remapping in two ways: to \textit{perturb} a model by finetuning it to respect a remapping, and to \textit{evaluate} a perturbed model by measuring the degree to which it respects a remapping.

\paragraph{Perturbation}
Let $p(S)$ be a language model and $M^{(T)}$ be a training remapping $\langle C_o^{(T)}, R_o^{(T)}, C_a^{(T)}, R_a^{(T)} \rangle$.
We perturb $p$ into $\tilde{p}_{M^{(T)}}$ by finetuning it (changing its weights) on $M^{(T)}$.
To reduce notational clutter, we omit the subscript on $\tilde{p}_{M^{(T)}}$ and superscripts on $C$ and $R$.
We minimize the following loss:
\begin{equation*}
\mathcal{L}\left(M^{(T)}, p\right) = \log {p\left(R_o | C_o\right)}  -\log p\left(R_a | C_a\right) 
\end{equation*}
In other words, we train the model to, e.g., use \textit{glam} instead of \textit{duck} in the context \textit{A \_\_\_ quacks}.

\paragraph{Evaluation}
We use remappings to evaluate the causal effect of perturbing $p$ into $\tilde{p}$.
Inspired by related work on causal interpretation of LMs \cite{arora-etal-2024-causalgym}, we take the log odds-ratio of the alternate to the original in an evaluation remapping 
$M^{(E)} = \langle C_o^{(E)}, R_o^{(E)}, C_a^{(E)}, R_a^{(E)}\rangle$, 
before and after perturbation (again omitting superscripts):

\begin{equation*}
\mathcal{R}\left(M^{(E)}, p, \tilde{p}\right) = \log\!\left[
\frac{\tilde{p}\left(R_a \mid C_a\right)}{p\left(R_a \mid C_a\right)}\cdot
\frac{p\left(R_o \mid C_o\right)} {\tilde{p}\left(R_o \mid C_o\right)}
\right]
\end{equation*}

For example, we can ask how perturbing the model on \perturb{A}{duck}{quacks}{glam} affects model behavior on \perturb{The}{duck}{led her ducklings}{glam}, thereby quantifying transfer to another instance of the same sense of \textit{duck}.
For simplicity of exposition, we define perturbation and evaluation above with respect to a single remapping, but both definitions can be applied without loss of generality to multiple remappings, simply by aggregating (e.g., averaging) over remappings.

\paragraph{Interpretation}
We impose no further formal constraints on remappings: the choice of remappings (via $C_o, R_o, C_a, R_a$) is left to the experimenter.
As with all science, appropriate experimental setup depends on the research question, and conclusions are only as strong as the experimental design on which they are based
(\citealt[][p. 18-26]{shadish2002experimental}; \citealt{zhang2024towards}).
For perturbation, which remappings are meaningful depends on the research question, and many remappings are likely nonsensical (e.g., \textit{A \textbf{duck} quacks} $\rightarrow$ \textit{Feed the \textbf{dog}}).
One way to create meaningful remappings---as we do here---is to fix the context (i.e., set $C_o$ and $C_a$ to be maximally similar; ideally identical) and vary the critical region.

\subsection{Experimental Design}
\label{sec:methods:exp}

\paragraph{Experiments}

We apply our method to three published datasets for linguistic abstraction learning across linguistic grain sizes.
\textit{First}, we study morphological representations via morphological categories in BATS \cite{gladkova-etal-2016-analogy}.
\textit{Second}, we study lexical representations via word sense disambiguation (WSD) in CoarseWSD-20 \cite{loureiro-etal-2021-analysis}.
\textit{Third}, we study syntactic representations via filler-gap dependency constructions in the dataset developed by \citet{boguraev-etal-2025-causal}.
We note that we use the terms \textit{morphology}, \textit{lexicon}, and \textit{syntax} merely as convenience labels, without implying mutual exclusivity \cite[see e.g.,][]{Bach1983-ww,Marantz1993-wv, goldberg1995}.

\paragraph{Implementation}
Because the perturbation objective $\mathcal{L}$ requires both $p\left(R_o | C_o \right)$ and $p\left(R_a | C_a \right)$, we run $S_a$ and $S_o$ through the model in a single batch, flip the sign of the gradients on $R_o$, and backpropagate in a single pass.
Models are trained using Adam \cite{kingma2014adam} with hyperparameter selection (learning rate and number of optimizer steps---generally 1-5) described in \aref{sec:app:hyper}.
Two of our benchmarks (morphological and lexical) require bidirectional context, so we use masked language models (MLMs): 
RoBERTa \cite{liu2019roberta}, 
modernBERT \cite{warner-etal-2025-smarter}, 
and GPT-BERT, a BabyLM trained on a developmentally plausible quantity of data \cite{warstadt-etal-2023-findings, charpentier-samuel-2024-bert}.
Our syntactic benchmark needs only preceding context, allowing us to use an autoregressive model, Pythia 1.4B \cite{pythia}, the same as used by \citet{boguraev-etal-2025-causal}.\footnote{
We obtain models from HuggingFace \cite{huggingface}.
}
In all cases, we equally weight the loss on all tokens of the critical regions in the two-entry batch.
For evaluation, since MLMs do not give $p(R | C)$ for multitoken words, we approximate it using the MLM pseudo-log-likelihood of \citet{kauf-ivanova-2023-better}.
All likelihoods are computed using the \texttt{minicons} library \cite{misra2022minicons}.\footnote{
\iffinal
    {Reproduction code and data are available at \repolink}
    {Reproduction code and data for all experiments is available at ANONYMIZED.}
}

\paragraph{Evaluation and metrics} 
Each of our benchmarks contains strings that implicitly belong to one or more latent classes (e.g., distinct senses of a word) whose representations we seek to study.
We use two types of evaluations.
\textit{First}, we quantify and visualize the effects of perturbation within and between classes, substantiating any claimed differences with regression-based statistical tests using the \margeff package \cite{marginaleffects}.
\textit{Second}, for the morphological and lexical benchmarks (where we have strong prior expectations that successful representation learning will yield discriminable latent classes), 
we use a generic notion of \textit{clusterability}, 
measuring the degree to which similarities are higher within class vs. between classes as an area under the curve (AUC, see \aref{app:methods:auc} for methodological details).

\paragraph{Controls}
As discussed in the \textbf{Introduction},
our primary concern is the scientific testing of psycholinguistic theories, by identifying causal representations in statistical learners (LMs).
For each experiment we thus use carefully designed controls.
Perturbation is unsupervised, in that it trains on only a single unlabeled training example.
This avoids, for example, issues relating to the kind of generalization licensed by a particular supervised task \cite{ hupkes2023taxonomy, jabbar-etal-2025-distinguishing}.
As the critical regions of perturbation and evaluation remappings are either identical (WSD) or disjoint (morphology, syntax), 
and as the minimal-pair evaluation setup holds contexts maximally fixed,
perturbation avoids confounds due to train--test target overlap \cite[see e.g., discussion of disjointness in][]{roznercryptics}. 
Robust control comparisons are a unique strength of perturbation, which avoids the \emph{faithfulness} concern of supervised approaches: 
that the analysis itself may learn the hypothesized structure rather than recover representations present in the model \cite{hewitt-liang-2019-designing,jacovi-goldberg-2020-towards}.
These aspects enable us to improve on the prior filler-gap work by \citet{boguraev-etal-2025-causal}, whose token-level confounds we both describe and avoid (see \sref{sec:fg}).

We further provide two \textbf{sanity checks}. 
We verify that single-example perturbation does not induce task learning: performance on the examples used for hyperparameter selection does not exceed performance on held-out test examples (\aref{sec:app:hyper}).
Additionally, for the syntax experiment, we find that under the selected hyperparameters, model perplexity degrades minimally under perturbation (\sref{sec:fg}).
\paragraph{Baselines}
Our experiments use datasets from established NLP tasks (e.g., WSD), enabling us to validate that perturbation recovers representational distinctions independently hypothesized in linguistics and psycholinguistics.

For both the morphology and WSD settings we further provide a cosine baseline as comparison (assessed via clusterability), to compare perturbation to established notions of (correlational) representations.
Recent work by \citet{teglia-etal-2025-much} found that the cosine similarity of contextualized word embeddings (especially intermediate layers) in MLMs reliably distinguished word senses.
Although this result aligns with linguistic expectations \cite[e.g.,][]{reif2019visualizing}, 
cosine similarity is correlational rather than causal
(see \sectionnameref{sec:intro}) and has been shown to suffer from a number of related drawbacks \cite{ethayarajh-2019-contextual, timkey-van-schijndel-2021-bark, zhou-etal-2022-problems, steck2024cosine}.
Because the \citeauthor{teglia-etal-2025-much} approach requires layer selection, we provide two baselines: last layer and best layer, which is
an oracle steel-man that gives a maximally conservative comparison.

\section{Morphological Representations}
\label{sec:morpho}

\newlength{\thwidth}
\setlength{\thwidth}{\dimexpr 0.325\textwidth-\tabcolsep}

\newcommand{\panel}[2]{%
\begin{overpic}[width=\linewidth]{#2}
    \put(-2,75){\sffamily \bfseries{#1}}
  \end{overpic}
}

\begin{table}[t]
\centering
\begin{NiceTabular}{l*{4}{m{0.8cm}}}
\toprule
\RowStyle{\bfseries}
Model & Perturb (trained) & Perturb (untrained) & Cos Best layer & Cos Last layer \\ 
\midrule
ModernBERT-base     & 0.837 & 0.529 & \textbf{0.912} & 0.772 \\
ModernBERT-large    & 0.857 & 0.537 & \textbf{0.883} & 0.722 \\
RoBERTa-base        & 0.867 & 0.514 & \textbf{0.868} & 0.764 \\
RoBERTa-large       & 0.889 & 0.518 & \textbf{0.893} & 0.657 \\
GPT-BERT            & 0.737 & 0.537 & \textbf{0.908} & 0.727 \\
\bottomrule
\end{NiceTabular}
\caption{
Morphology: Average AUC clusterability.
}
\label{tab:morpho}
\end{table}

\begin{figure*}
\begin{tabular}{@{}p{\thwidth}p{\thwidth}p{\thwidth}@{}}
    \panel{A}{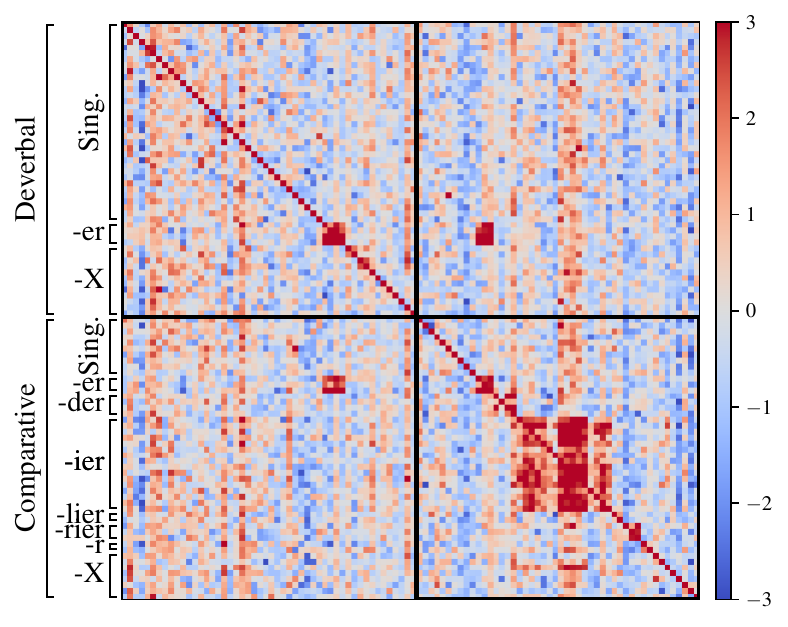} &
    \panel{B}{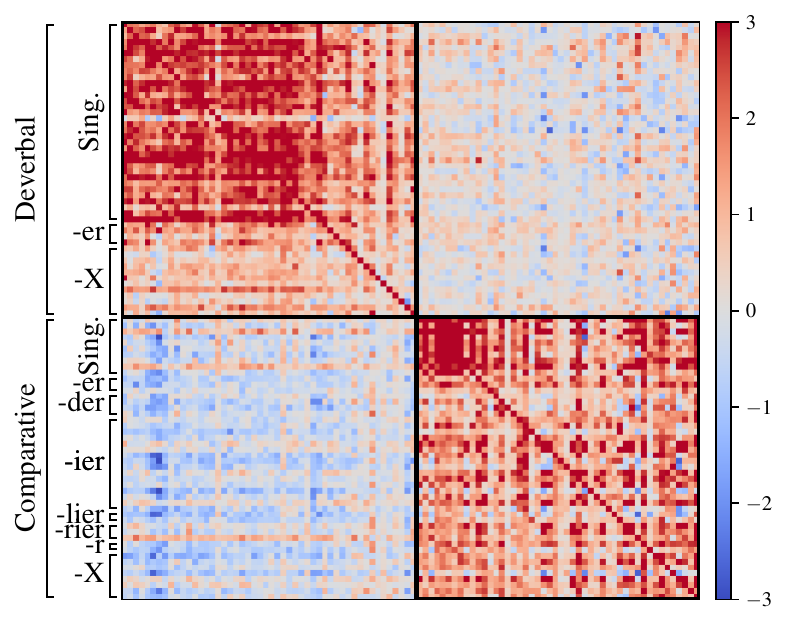} &
    \raisebox{-14pt}{
        \begin{overpic}[width=\linewidth]{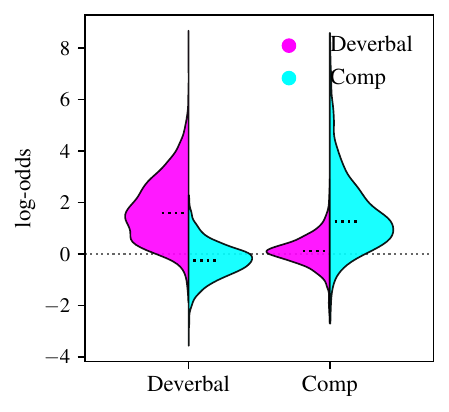}
            \put(-2,85){\sffamily \bfseries{C}}
          \end{overpic}
    }
\end{tabular}
\vspace{-10pt}
\caption{
\textbf{A}, \textbf{B}: Transfer for Morphology task in RoBERTa-large. 
\textbf{A} is randomly initialized; 
\textbf{B} is the trained model.
Cell $i, j$ corresponds to finetuning on the $i$\textsuperscript{th} remapping and evaluating on the $j$\textsuperscript{th}.
Items are grouped hierarchically by morphological class (outer, deverbal vs. comparative) and tokenization (inner, e.g., \textit{Sing.} means single token, \textit{-er} means the final token is \textit{-er}, etc.).
\textbf{C}: Distribution of transfer effects (log-odds) to deverbal (magenta) vs. comparative (cyan) depending on which class is perturbed ($x$-axis), with the median indicated in each distribution.
}
\label{fig_morph}
\end{figure*}

\noindent\textbf{Methods:} We ask whether perturbation distinguishes two uses of the suffix \emph{-er}: deverbal (e.g., \ttw{teach} vs. \ttw{teacher}) and comparative (e.g., \ttw{tall} vs. \ttw{taller}).
We start with the BATS dataset \cite{gladkova-etal-2016-analogy}, which has 50 pairs for both classes, and obtain naturalistic usages by sampling sentences from the GPT-BERT Cosmo corpus (\citealt{charpentier-samuel-2024-bert}).
We create remappings for perturbation and evaluation by removing the \emph{-er} suffix:
For example:
\perturb{The man was}{taller}{than the boy}{tall}.

\noindent\textbf{Results:} We report results for RoBERTa-large on 98 sentences.
We find (\figref{fig_morph}{A}) that in untrained models, transfer occurs only due to shared tokens;\footnote{
Vertical patterning is observed in untrained models, indicating that at initialization, some items are more likely to be influenced than others overall. 
Permutation tests show that this patterning is unpredictable across initializations (\aref{app:morpho:permutations}).
}
By contrast, in trained models (\figref{fig_morph}{B}), transfer is instead dominated by class membership, as is clear in aggregate (\figref{fig_morph}{C}).
Representations from RoBERTa-large give clusterability of 0.89; representations at random initialization give clusterability of 0.52, which is close to chance, as expected.
Perturbation thus qualitatively shows the emergence of a clear deverbal-comparative distinction during training.

In \autoref{tab:morpho} we compare perturbation to the correlational cosine baseline. 
Perturbation outperforms last-layer cosine, but the oracle best-layer cosine is better in all models.
As there are significant effects of tokenization on transfer (see below), we further segment performance by tokenization in \aref{sec:app:morph:cos}.
In the single-token setting, perturbation outperforms the best-layer oracle baseline on 4 of 5 models, even when the latter is allowed to ``peek'' at its own performance in choosing which layer to use.
This suggests that future work should look to improve the perturbation methodology for the multitoken setting.
(The multitoken setting for morphology is complicated for perturbation, since we are often perturbing a multitoken word like ``merrier'' to a single-token one like ``merry,'' giving a two-entry batch that is not aligned.)
Although these results show that cosine similarity can separate distinct morphological senses, cosine similarity does not establish a causal role of the representations (see \sref{sec:methods});
we include the cosine similarity baseline only as a reference for comparison.

We test the perturbation effect quantitatively using linear regression with three factors: 
\textit{class} has three levels: within-class comparative, within-class deverbal, and between-class;
\textit{trained} is true or false;
\textit{tokenization} has four levels: single-single (the critical regions of both the perturbation and evaluation are single-token), single-multi (one of the critical regions is single-token and one is multi-token), multi-different (both regions are multitoken with differing final tokens),
and multi-same (both regions are multitoken with the same final token).
The response variable is the cell values (log-odds ratio) from the matrices in \figref{fig_morph}{A-B}, omitting the diagonal and symmetrizing by averaging the two interactions for each pair (9,506 observations).

We find that trained models show significant within-class transfer (1.73, $p<0.001$) 
and minimal between-class transfer (0.003, $p=0.97$; $\Delta=1.72, p<0.001$).
By contrast, untrained models show comparatively weaker within class transfer (0.56, $p<0.001$) and stronger between-class transfer (0.75, $p<0.001$).
The interaction between \textit{class} and \textit{training} is significant: within class transfer is larger in trained than untrained models ($\Delta=1.17, p<0.001$); and between-class transfer is smaller ($\Delta=-0.75, p< 0.001$).
Moreover, the significant transfer both within and between classes in untrained models is driven by shared tokenization: average transfer when a token is shared is 2.6 for within-class and 3.3 for between-class transfer, whereas when there is no shared token, these values drop to \mbox{-0.12} and \mbox{-0.08} respectively
(both differences are significant, $p<0.001$).
By contrast, for trained models, although some tokenization effects persist, average between-class transfer with shared tokenization is only 0.04 (vs. untrained $=-3.2, p<0.001$),
showing that, over the course of training, RoBERTa learns to ignore superficially shared tokens and prioritize the representation of the word as a whole.
See \aref{sec:app:morph:lm} for full details and a tabular presentation of the linear model.
In \aref{app:morpho:addnal} we further investigate which aspects of the perturbation remapping setup contribute to transfer.

\section{Lexical Representations}
\label{sec:wsd}

\begin{table}[t]
\centering
\begin{NiceTabular}{l*{4}{m{0.8cm}}}
\toprule
\RowStyle{\bfseries}
Model & Perturb (trained) & Perturb (untrained) & Cos Best layer & Cos Last layer \\ 
\midrule
modernBERT-B & \textbf{0.829} & 0.556 & 0.755 & 0.738 \\ 
modernBERT-L & \textbf{0.871} & 0.553 & 0.801 & 0.769 \\ 
RoBERTa-B & \textbf{0.811} & 0.502 & 0.798 & 0.759 \\ 
RoBERTa-L & 0.845  & 0.515 & \textbf{0.889}& 0.816 \\ 
GPT-BERT & 0.692 & 0.563 & \textbf{0.713} & 0.693 \\ 

\bottomrule
\end{NiceTabular}
\caption{
Lexical Representations: Average AUC clusterability over all 20 words from CWSD. 
}
\label{tab:wsd}
\end{table}

\begin{figure}[t]
\includegraphics[width=\linewidth]{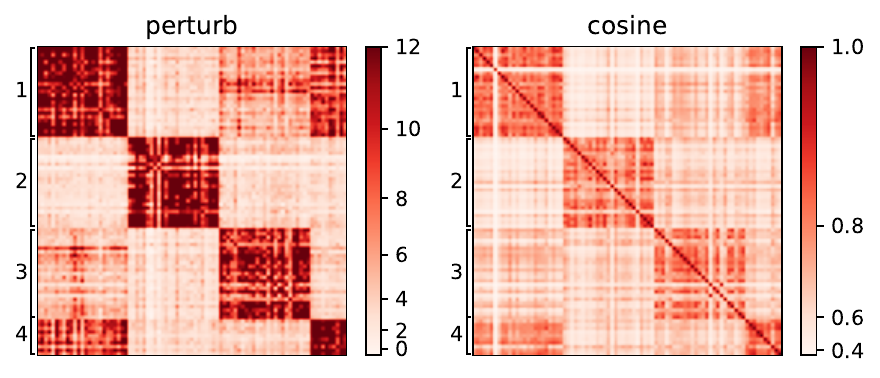}
\vspace{-15pt}
\caption{
Similarity matrices for \emph{square} (WSD) using perturbation (left) and cos last (right).
Block structure corresponds to four senses: shape, company, town square, numerical square (e.g., 4,9).
AUCs: 0.94, 0.87
}
\label{fig:wsd:square_main}
\end{figure}

\noindent\textbf{Methods:} We ask whether perturbation reveals latent distinctions between different senses for the same word string (word sense disambiguation, WSD), using the CoarseWSD-20 dataset (CWSD; \citealt{loureiro-etal-2021-analysis}).
CWSD is a coarse-grained, naturalistic dataset drawn from Wikipedia with 20 nouns, each with two to five distinct senses labeled by expert annotators.
In general, the intended word sense is strongly indicated by the sentential context.
In all cases, we perturb the word of interest to the nonce token \textit{glam} (e.g., \perturb{He is eating an}{apple}{}{glam}),
though other nonce tokens yield qualitatively similar results (\aref{app:wsd:nonce}).

For each of the 20 words, we sample 100 examples equally balanced over word senses (see \aref{app:wsd:methods}).
The similarity measures are evaluated against ground-truth labels using the clusterability technique introduced in \sref{sec:methods:exp};
we compute clusterability for each word and then average over all words.

\noindent\textbf{Results:} \autoref{fig:wsd:square_main} provides a  \mbox{modernBERT-L} similarity matrices for senses of \emph{square} using perturbation and cosine similarity.
Exhaustive visualizations, including for untrained models, are given in \aref{app:wsd:wsd_viz}.
As shown, both methods reveal latent representations for distinct word senses, but this influence tends to be more pronounced under perturbation, suggesting that the causal influence of word sense representations in LMs may be even larger than the correlational impression given by state similarities.
We evaluate this impression quantitatively in \autoref{tab:wsd}, where perturbation clusterability is higher than the oracle best-layer clusterability in three of five models. 
Our results therefore converge with prior findings that 
LMs represent word senses \cite{teglia-etal-2025-much},
while also grounding this conclusion in causal effects on model learning.

\begin{figure*}
\includegraphics[width=\linewidth]{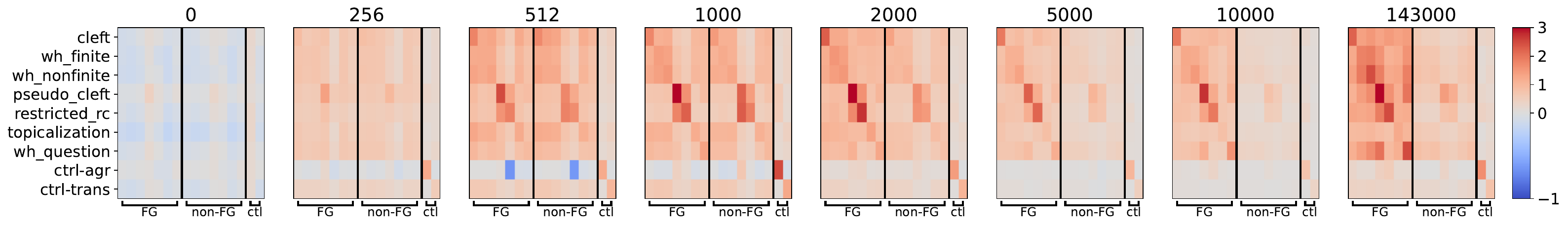}
\vspace{-20pt}
\caption{Transfer (aggregated over animacy and embeddedness conditions) \emph{from} each FG and control \emph{to} FG, (minimal) non-FG conditions, and controls.
FG and non-FG exhibit similar transfer through step 1k, after which transfer to non-FG begins decreasing.
Once FG is learned (step 2k-5k), minimal transfer is seen to/from controls.
}
\label{fig_4}
\end{figure*}

\begin{figure}[t]

\begin{overpic}[width=\linewidth]{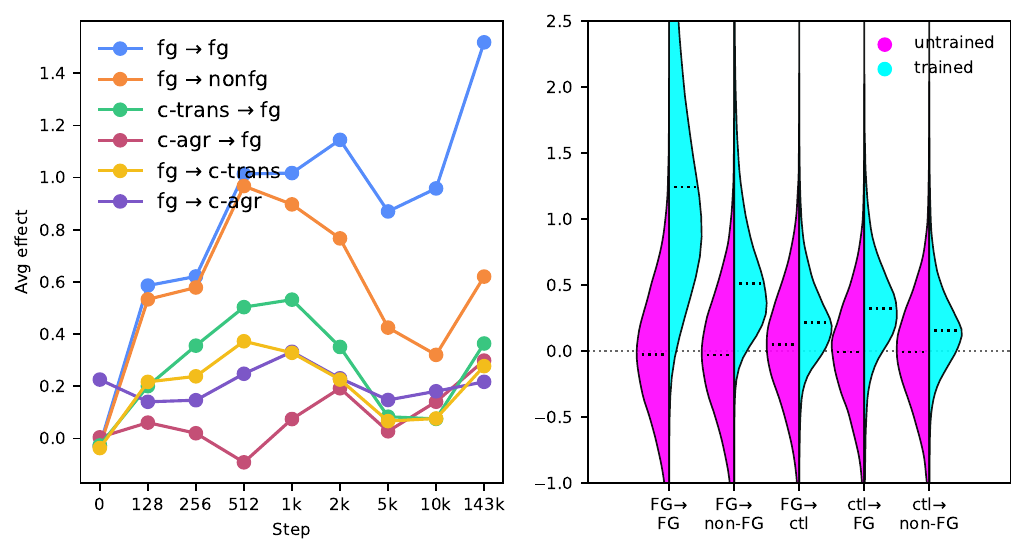}
    \put(2,48){\sffamily \bfseries{A}}
    \put(49.5,48){\sffamily \bfseries{B}}
  \end{overpic}
\vspace{-20pt}
\caption{
\textbf{A}: Transfer between indicated classes over training. 
Std errors are too small to be seen.
\textbf{B}: Transfer between indicated classes in untrained (magenta) and trained (cyan) models, with medians shown.
}
\label{fig:fg_aggs}
\end{figure}

\section{Syntactic Representations}
\label{sec:fg}
\noindent\textbf{Methods:}
We turn finally to LM representations of syntax via filler-gap (FG) constructions \cite[e.g., \emph{I know who$_i$ the man liked [\textit{gap}$_i$]}, where the referent of \emph{who} is indexed to the referent of a phonologically null \textit{gap} in its source position;][]{Ross1967-je}.
Diverse types of FG constructions exist (e.g., English has \textit{wh}-questions, relative clauses, clefts and pseudoclefts, among others), and linguists debate the degree of shared representation between them \cite[e.g.,][]{Ross1967-je,Chomsky1986-bu,Culicover1999-pf,Levine2006-bq}.
Thus, unlike coarse-grained word senses that enjoy relatively strong consensus, the expected representation of FG constructions in English is an open question, one which perturbation allows us to study through the vehicle of LMs.
Recently, \citet{boguraev-etal-2025-causal} showed that DAS finds representational transfer in LMs across diverse filler-gap constructions.
But given open concerns about DAS (i.e., the risk of reifying researcher assumptions and/or missing nonlinear representations, see \sectionnameref{sec:intro}), here we revisit this finding using perturbation.

The \citeauthor{boguraev-etal-2025-causal} dataset contains seven FG types: two embedded question types (\textit{know} vs.\ \textit{wonder}), matrix \textit{wh}-questions, restrictive relatives, clefts, pseudoclefts, and topicalization (see \aref{app:fg}, \autoref{tab:app:bog_tab} for details).
For each type, the authors designed pairs of sentences that vary a minimal part of the context with a corresponding change in output target (e.g., \minp{I know}{who}{that}{the man liked}{.}{him}).
These pairs are also varied along two dimensions: \textit{animacy} of the extracted element (e.g., \textit{I know \textbf{[what, who]} the man saw}) and \textit{embeddedness} of the extraction site (e.g., \textit{I know who \textbf{[the nurse said that]} the man liked}).
When studied using DAS, the design above ostensibly measures sensitivity to relatedness between FG constructions; it does not measure selectivity to FG over non-FG constructions.
To address this, \citeauthor{boguraev-etal-2025-causal} added two control conditions to serve as foils: (a) an agreement control (\ctla) that manipulates subject-verb agreement
(\minp{The}{boy}{boys}{the man liked}{is}{are})
and 
(b) a transitivity control (\ctlt) that manipulates the transitivity of the verb (\minp{That man}{ran}{liked}{}{.}{him}).

This dataset provides natural remappings for perturbation and evaluation.
For example, we can perturb on \perturb{I know who the man liked}{.}{}{him} and evaluate transfer on \perturb{Who the man liked}{was}{}{it}.
Conceptually, this allows us to trace corruptions of the model's native filler-gap representations.
However, perturbation offers a key advantage over DAS for this benchmark: whereas DAS needs both the FG and non-FG condition 
(e.g., \textit{I know \textbf{that} the man liked \textbf{[., him]}}) 
to train its intervention 
(since activations from both items in a pair are used for each intervention),
perturbation needs only the FG condition.
Thus with perturbation, we can use the non-FG condition as an additional,  minimally different control.
We consider non-FG a much stronger control than \ctla and \ctlt used by \citeauthor{boguraev-etal-2025-causal}, and
while we report transfer to \ctla and \ctlt, we focus on the minimal non-FG controls in our interpretation of results.
We also make our train and evaluation mappings disjoint with respect to the output target: 
all evaluation remappings have critical regions that differ from the critical regions of the perturbation remapping, thus enabling a more conservative test of transfer than the one used by \citeauthor{boguraev-etal-2025-causal} (which does not have the same disjointness property).

The different FG conditions differ not only in construction type but also in which words can sensibly occupy the critical region.
For example, the embedded question conditions might contrast punctuation with a pronoun, whereas pseudoclefts might contrast a verb with a pronoun (see \autoref{tab:app:bog_tab}).
Similarity of causal transfer (using either DAS or perturbation) will plausibly be influenced by similarity (both exact lexical or, e.g., part of speech) of evaluation targets, in addition to the influence of any abstractions (e.g., FG) they might share.
We substantiate this concern in \aref{app:fg:jaccard_sim}, where we show that there are large differences in \emph{critical region similarity} (\crshort)---the degree of lexical overlap between possible completions (critical regions) for different pairs of FG conditions in the \citeauthor{boguraev-etal-2025-causal} dataset.
DAS cannot directly control for this confound, which creates a previously undocumented interpretational problem for their results.\footnote{
Note that exact lexical match between training and evaluation completions may also pose a problem, since \citeauthor{boguraev-etal-2025-causal} did not use disjoint targets when training DAS.
Given that DAS can cause models to output arbitrary targets  \cite{wu2023interpretability, arora-etal-2024-causalgym} this may further confound their results.
Our evaluation eliminates this issue.
}
For example, \ctla has almost no \crshort with any FG condition (\autoref{fig:app:fg_label_sim_fg}), so it is an anticonservative control.
By contrast, we can and do control for this confound using perturbation, simply by subtracting out the transfer to the minimal non-FG control mentioned above (which has maximally similar context and evaluates against identical critical regions).
This yields a tightly controlled estimate of the portion of perturbation transfer that is selectively driven by FG, over and above superficial contextual and lexical similarities.
Henceforth we refer to this controlled estimate (i.e., the pairwise difference between transfer in FG$\rightarrow$FG and FG$\rightarrow$non-FG) as \blt because it subtracts out the baseline transfer rate.

\noindent\textbf{Results:} 
Perturbation transfer across Pythia 1.4B training checkpoints is shown in \autoref{fig_4} from each FG condition and from the \citeauthor{boguraev-etal-2025-causal} control conditions (rows) to 
(i) other FG conditions (left columns), 
(ii) minimal non-FG condition (middle columns), and 
(iii) \ctla and \ctlt (right columns),
aggregating for simplicity across the animacy and embeddedness conditions. 
\Blt is obtained by subtracting the non-FG block from the FG block.
We provide unaggregated and \blt versions in \autoref{fig:app:fg_acq_all}.
As shown, the model initially shows little between-condition transfer (step 0), but quickly acquires transfer primarily as a function of \crshort.
For example, at step 512, transfer is highly similar between the FG and non-FG conditions, with especially strong transfer between the restrictive relative and pseudocleft conditions, which have common targets (\aref{app:fg:jaccard_sim}, \autoref{fig:app:fg_label_sim_fg}).
By step 2k, however, we see the model begin to distinguish FG from the minimal non-FG controls, and transfer is much stronger to FG over non-FG conditions by the end of training.
\figref{fig:fg_aggs}{A} supports this general impression by showing aggregate transfer over the course of training:
FG perturbations transfer roughly equally to FG and non-FG until step 512, when they begin to strongly diverge.
Models show some transfer to \ctlt initially, which peaks around step 512, begins to decrease at step 1k---which is also when the model begins to distinguish FG from non-FG---and mostly vanishes by step 5k. 
We speculate on causes of this behavior in \aref{app:fg:disc}.
Our results reveal not only the structure of representations at specific checkpoints but also the dynamics of their emergence over training, both of which show interpretable patterns that align with linguistic theorizing.

We also follow up on \citeauthor{boguraev-etal-2025-causal}'s report of minimal transfer between embedded and non-embedded FGs.
Our approach is more conservative than theirs
(we use fully disjoint training and evaluation sets and intervene unsupervised from a single training example), but we
nonetheless find substantial \blt between the embeddedness conditions (see \aref{app:fg}, \autoref{fig:app:fg_acq_all}, right column, especially steps 2k, 5k, and 10k, where between-condition structure begins to emerge; statistical tests of significance in \autoref{app:fg:stat}), 
suggesting some degree of common representation between FG constructions even when composed with unrelated constructions (subordinate clauses).

In \figref{fig:fg_aggs}{B} we plot distributions of (unbaselined) transfer from FG perturbations to three broad classes of examples (other FG conditions, non-FG conditions, and \citeauthor{boguraev-etal-2025-causal} controls); as well as from the \citeauthor{boguraev-etal-2025-causal} controls to the FG and non-FG conditions.
Untrained models show no transfer across the board.
Trained models generalize FG perturbations most strongly to other FG conditions, with much weaker transfer in the other comparisons (see \aref{app:fg:stat} for statistical comparisons). 
In \aref{app:fg:fg_sims}, we briefly discuss gradations in similarity among the individual FG conditions.
Finally, as a sanity check, we measure the change in perplexity under FG perturbations on a wikitext sample (10k words; \citealt{merity2017pointer}):
Pythia-1.4B perplexity changes minimally from 23.58 to 23.60, suggesting that perturbation has fairly selective effects and largely preserves overall model function.

\section{Discussion}
\label{sec:disc}
We developed perturbation as a bottom-up method for studying LM representations, based on the hypothesis that representations are conduits for learning \cite{hinton1986learning, shepard1987toward}.
We supported this hypothesis by showing that perturbation reveals interpretable morphological, lexical, and syntactic structure.

Perturbation has several advantages over other techniques.
\textit{First}, perturbation is data-driven: it uses only a single unlabeled training example, and we provide no guidance to the model about how we expect it to generalize the perturbation.
Perturbation thus avoids the interpretational pitfalls of supervised interpretability methods like probing and DAS---including the risk of reifying researcher assumptions (see \sectionnameref{sec:intro}).

\textit{Second}, this data-drivenness allows the experimenter to make inferences not only about whether a model contains a hypothesized representation, but also about how functionally important that representation is.
For example, in the filler-gap analysis, we observe transfer not only along the filler-gap dimension, but also according to animacy and embeddedness, which were not explicit objects of study.

\textit{Third}, perturbation makes no geometric assumptions, so it can recover arbitrary representations.

\textit{Fourth}, we have shown empirically that, unlike supervised approaches like DAS \cite{arora-etal-2024-causalgym}, perturbation does not find high-level representations in untrained models.
Selectivity for real over spurious representations has long been a sought-after (and hard-to-achieve) property of interpretability techniques \cite{hewitt-liang-2019-designing,arora-etal-2024-causalgym, sutter2025non}, and we find that perturbation is naturally \emph{faithful}.
Moreover, because perturbation transfer is weak when representations are absent, perturbation can support nuanced inferences about the emergence of representations during training, with potential implications both for pretraining design and for studying statistical learning in humans via LM proxies \cite{warstadt2022artificial, frank2023bridging}.
For example, our perturbation-based findings lend new support to claims that usage is sufficient to learn much morphological, lexical, and syntactic structure \cite{goldberg2003constructions,tomasello2005constructing,bybee2006}.

\textit{Fifth},
perturbation's simplicity makes it both data- and compute-efficient (\aref{sec:app:methods:cost}), requiring only one training instance per intervention, a small number (1-5) of gradient steps for training, and a single forward pass per evaluation batch.
This contrasts with supervised approaches like probing and DAS, which require sufficient training data to train additional projections of model states, often at each layer in the model.
Perturbation is thus compatible with small naturalistic datasets.

\textit{Sixth}, as shown in \sref{sec:fg}, since perturbation requires only a single remapping, it enables robust experimental controls that are not possible with DAS, which necessarily needs the control remapping for both training and evaluating interventions.

Lastly, perturbation is a simple and inexpensive tool for linguists studying the relatedness of linguistic phenomena \cite[e.g.,][]{goldberg1995, stefanowitsch2003, langacker2005construction, bybee2010language, diessel2023constructicon}.
One recently developed method for this purpose is corpus-ablation-based acquisition studies \cite{misra-mahowald-2024-language, patil-etal-2024-filtered, leong-linzen-passive}.
Though these methods have been successful, they require meticulous curation of relatively large corpora and potentially days-long retraining of even small LMs.

\section{Conclusion}
We introduced perturbation, a simple, data-driven method for studying representations in language models. 
Across morphological, lexical, and syntactic structures, perturbation systematically revealed interpretable linguistic representations that emerge with training.
Our work offers an escape from the dilemma faced by supervised methods between being too expressive and not expressive enough, while also converging with and strengthening prior claims that many linguistic abstractions can be acquired from distributional evidence alone.

\section*{Limitations}
\label{sec:lim}
This work has several limitations.
\textit{First}, as discussed (\sectionnameref{sec:methods}), although perturbation is unsupervised in the sense that it trains on only a single unlabeled example,
the interpretability of results is only as good as the selection of remappings and controls.
As with most interpretability methods, researchers must take care to set up the experiment to license meaningful conclusions (e.g., \citealt[][p. 18-26]{shadish2002experimental}; \citealt{zhang2024towards}).

\textit{Second}, we found that perturbation transfer was influenced by tokenization (\sref{sec:morpho}).
Although perturbation clusterability is higher than oracle best-layer cosine for single-token words, clusterability in multitoken words is substantially lower (\sref{sec:app:morph:cos}).
This may be an inherent challenge in perturbing a multitoken word to a single token word, as was often the case when perturbing a word like ``merrier'' to ``merry''. 

\textit{Third}, although we tested perturbation on both autoregressive and masked language models, our current implementation does not support perturbing autoregressive models for tasks that use bidirectional context.
For example, we would not be able to reliably study word sense disambiguation in autoregressive models, because disambiguating sense information often follows the ambiguous word.
This limitation results from the structure of the task, and other interpretability methods like DAS would have the same problem. 
Given the prevalence of autoregressive models in actual use, future work should focus on addressing this limitation, perhaps via an approach like echo embeddings \cite{springer2025repetition}.

\textit{Fourth}, we focused for practical reasons exclusively on linguistic representations in English.
Perturbation is perfectly amenable to studying representations in other languages as well as representations of nonlinguistic knowledge (e.g., commonsense, social, or situational knowledge).
We hope our proposal will support future research along these lines.

\textit{Fifth}, unlike approaches like DAS \cite{geiger2024finding} or circuit analysis \cite{Elhage2021-qh}, perturbation is more useful for finding shared representations than controllable ``mechanisms.''
For example, although perturbation can detect when LMs distinguish different senses of a given word, it does not reveal how the senses are encoded by the model or how to intervene on that encoding in order to steer model behavior.
Perturbation may ultimately be useful as an initial guide to identifying steerable mechanisms under weaker geometrical constraints than e.g., DAS, but more work is needed to explore this possibility.

In this work, because our goal was to establish perturbation as a faithful method for studying linguistic abstractions, we used a simple fine-tuning setup in which all model parameters were allowed to vary.
Future work should compare perturbation under parameter-efficient fine-tuning \cite{houlsby-peft,hu2022lora} or with parameter updates restricted to particular modules, or layers 
\cite{akbartajari-etal-2022-empirical}. 
Such restrictions could improve the computational scalability of perturbation and also enable more targeted hypotheses about where and how linguistic abstractions are represented in the network.

\textit{Sixth}, although our work was partially motivated by usage-based theories of grammar, we recognize that the recovery of linguistic structure in statistical learners with no innate symbolically-specified grammar (LMs) is not necessarily incompatible with the existence of innate grammatical constraints in humans.
Our work instead contributes to a growing body of evidence on the feasibility of bottom-up statistical learning of linguistic structure, which is one of many dimensions of debate on the human capacity for language.

\iffinal{
\section*{Acknowledgements}
We are grateful to Leonie Weissweiler, Kyle Mahowald, Emily Goodwin, Jasper Jian, Nathan Roll, Yuhan Zhang, Amir Zur, and other members of the CLiMB Lab at Stanford for comments and discussion.
We thank Sasha Boguraev for answering questions about the \citet{boguraev-etal-2025-causal} methods and dataset.

} 

\bibliography{anthology-1, anthology-2, josh, literatur_25_03, josh_bib_local, shain}

\begin{thebibliography}{111}
\providecommand{\natexlab}[1]{#1}

\bibitem[{Adi et~al.(2017)Adi, Kermany, Belinkov, Lavi, and Goldberg}]{adi2017finegrained}
Yossi Adi, Einat Kermany, Yonatan Belinkov, Ofer Lavi, and Yoav Goldberg. 2017.
\newblock \href {https://openreview.net/forum?id=BJh6Ztuxl} {Fine-grained analysis of sentence embeddings using auxiliary prediction tasks}.
\newblock In \emph{International Conference on Learning Representations}.

\bibitem[{AkbarTajari et~al.(2022)AkbarTajari, Rajaee, and Pilehvar}]{akbartajari-etal-2022-empirical}
Mohammad AkbarTajari, Sara Rajaee, and Mohammad~Taher Pilehvar. 2022.
\newblock \href {https://doi.org/10.18653/v1/2022.emnlp-main.726} {An empirical study on the transferability of transformer modules in parameter-efficient fine-tuning}.
\newblock In \emph{Proceedings of the 2022 Conference on Empirical Methods in Natural Language Processing}, pages 10617--10625, Abu Dhabi, United Arab Emirates. Association for Computational Linguistics.

\bibitem[{Alain and Bengio(2017)}]{alain2017understanding}
Guillaume Alain and Yoshua Bengio. 2017.
\newblock \href {https://openreview.net/forum?id=HJ4-rAVtl} {Understanding intermediate layers using linear classifier probes}.
\newblock ICLR 2017 Workshop Track.

\bibitem[{Arel-Bundock et~al.(2024)Arel-Bundock, Greifer, and Heiss}]{marginaleffects}
Vincent Arel-Bundock, Noah Greifer, and Andrew Heiss. 2024.
\newblock \href {https://doi.org/10.18637/jss.v111.i09} {How to intrpret statistical models using {marginaleffects} for {R} and {Python}}.
\newblock \emph{Journal of Statistical Software}, 111(9):1--32.

\bibitem[{Arora et~al.(2024)Arora, Jurafsky, and Potts}]{arora-etal-2024-causalgym}
Aryaman Arora, Dan Jurafsky, and Christopher Potts. 2024.
\newblock \href {https://doi.org/10.18653/v1/2024.acl-long.785} {{C}ausal{G}ym: Benchmarking causal interpretability methods on linguistic tasks}.
\newblock In \emph{Proceedings of the 62nd Annual Meeting of the Association for Computational Linguistics (Volume 1: Long Papers)}, pages 14638--14663, Bangkok, Thailand. Association for Computational Linguistics.

\bibitem[{Bach(1983)}]{Bach1983-ww}
Emmon Bach. 1983.
\newblock On the relationship between word-grammar and phrase-grammar.
\newblock \emph{Nat. Lang. Linguist. Theory}, 1(1):65--89.

\bibitem[{Belinkov(2022)}]{belinkov-2022-probing}
Yonatan Belinkov. 2022.
\newblock \href {https://doi.org/10.1162/coli_a_00422} {Probing classifiers: Promises, shortcomings, and advances}.
\newblock \emph{Computational Linguistics}, 48(1):207--219.

\bibitem[{Biderman et~al.(2023)Biderman, Schoelkopf, Anthony, Bradley, O'Brien, Hallahan, Khan, Purohit, Prashanth, Raff, Skowron, Sutawika, and Van Der~Wal}]{pythia}
Stella Biderman, Hailey Schoelkopf, Quentin Anthony, Herbie Bradley, Kyle O'Brien, Eric Hallahan, Mohammad~Aflah Khan, Shivanshu Purohit, USVSN~Sai Prashanth, Edward Raff, Aviya Skowron, Lintang Sutawika, and Oskar Van Der~Wal. 2023.
\newblock Pythia: a suite for analyzing large language models across training and scaling.
\newblock In \emph{Proceedings of the 40th International Conference on Machine Learning}, ICML'23. JMLR.org.

\bibitem[{Boguraev and Mahowald(2026)}]{boguraev2026drawbridge}
Sasha Boguraev and Kyle Mahowald. 2026.
\newblock \href {https://arxiv.org/abs/2604.13950} {Causal drawbridges: Characterizing gradient blocking of syntactic islands in transformer lms}.
\newblock \emph{Preprint}, arXiv:2604.13950.

\bibitem[{Boguraev et~al.(2025)Boguraev, Potts, and Mahowald}]{boguraev-etal-2025-causal}
Sasha Boguraev, Christopher Potts, and Kyle Mahowald. 2025.
\newblock \href {https://doi.org/10.18653/v1/2025.emnlp-main.1271} {Causal interventions reveal shared structure across {E}nglish filler{--}gap constructions}.
\newblock In \emph{Proceedings of the 2025 Conference on Empirical Methods in Natural Language Processing}, pages 25021--25042, Suzhou, China. Association for Computational Linguistics.

\bibitem[{Branigan and Pickering(2017)}]{branigan-pickering-2017}
Holly~P. Branigan and Martin~J. Pickering. 2017.
\newblock \href {https://doi.org/10.1017/S0140525X16002028} {An experimental approach to linguistic representation}.
\newblock \emph{Behavioral and Brain Sciences}, 40:e282.

\bibitem[{Bybee(2006)}]{bybee2006}
Joan Bybee. 2006.
\newblock From {U}sage to {G}rammar: {T}he {M}ind’s {R}esponse to {R}epetition.
\newblock \emph{Language}, 82(4):711--733.

\bibitem[{Bybee(2010)}]{bybee2010language}
Joan Bybee. 2010.
\newblock \emph{Language, usage and cognition}.
\newblock Cambridge University Press.

\bibitem[{Charpentier et~al.(2025)Charpentier, Choshen, Cotterell, Gul, Hu, Liu, Jumelet, Linzen, Mueller, Ross, Shah, Warstadt, Wilcox, and Williams}]{charpentier-etal-2025-findings}
Lucas Charpentier, Leshem Choshen, Ryan Cotterell, Mustafa~Omer Gul, Michael~Y. Hu, Jing Liu, Jaap Jumelet, Tal Linzen, Aaron Mueller, Candace Ross, Raj~Sanjay Shah, Alex Warstadt, Ethan~Gotlieb Wilcox, and Adina Williams. 2025.
\newblock \href {https://doi.org/10.18653/v1/2025.babylm-main.28} {Findings of the third {B}aby{LM} challenge: Accelerating language modeling research with cognitively plausible data}.
\newblock In \emph{Proceedings of the First BabyLM Workshop}, pages 399--420, Suzhou, China. Association for Computational Linguistics.

\bibitem[{Charpentier and Samuel(2024)}]{charpentier-samuel-2024-bert}
Lucas Georges~Gabriel Charpentier and David Samuel. 2024.
\newblock \href {https://aclanthology.org/2024.conll-babylm.24/} {{GPT} or {BERT}: why not both?}
\newblock In \emph{The 2nd BabyLM Challenge at the 28th Conference on Computational Natural Language Learning}, pages 262--283, Miami, FL, USA. Association for Computational Linguistics.

\bibitem[{Chomsky(1986)}]{Chomsky1986-bu}
Noam Chomsky. 1986.
\newblock \emph{Barriers}.
\newblock The MIT Press, Massachusetts Institute of Technology, Cambridge.

\bibitem[{Conneau et~al.(2018)Conneau, Kruszewski, Lample, Barrault, and Baroni}]{conneau-etal-2018-cram}
Alexis Conneau, German Kruszewski, Guillaume Lample, Lo{\"i}c Barrault, and Marco Baroni. 2018.
\newblock \href {https://doi.org/10.18653/v1/P18-1198} {What you can cram into a single {\$}{\&}!{\#}* vector: Probing sentence embeddings for linguistic properties}.
\newblock In \emph{Proceedings of the 56th Annual Meeting of the Association for Computational Linguistics (Volume 1: Long Papers)}, pages 2126--2136, Melbourne, Australia. Association for Computational Linguistics.

\bibitem[{Csord{\'a}s et~al.(2024)Csord{\'a}s, Potts, Manning, and Geiger}]{csordas-etal-2024-recurrent}
R{\'o}bert Csord{\'a}s, Christopher Potts, Christopher~D Manning, and Atticus Geiger. 2024.
\newblock \href {https://doi.org/10.18653/v1/2024.blackboxnlp-1.17} {Recurrent neural networks learn to store and generate sequences using non-linear representations}.
\newblock In \emph{Proceedings of the 7th BlackboxNLP Workshop: Analyzing and Interpreting Neural Networks for NLP}, pages 248--262, Miami, Florida, US. Association for Computational Linguistics.

\bibitem[{Culicover(1999)}]{Culicover1999-pf}
Peter~W Culicover. 1999.
\newblock \emph{Syntactic Nuts: Hard Cases, Syntactic Theory, and Language Acquisition}.
\newblock Oxford University Press, London, England.

\bibitem[{Desai and Nair(2026)}]{desai-nair-2026-filling}
Atrey Desai and Sathvik Nair. 2026.
\newblock \href {https://doi.org/10.18653/v1/2026.findings-acl.1737} {Filling in the mechanisms: How do {LM}s learn filler-gap dependencies under developmental constraints?}
\newblock In \emph{Findings of the {A}ssociation for {C}omputational {L}inguistics: {ACL} 2026}, pages 34799--34815, San Diego, California, United States. Association for Computational Linguistics.

\bibitem[{Diego~Simon et~al.(2024)Diego~Simon, d'Ascoli, Chemla, Lakretz, and King}]{diego2024polar}
Pablo~J Diego~Simon, St{\'e}phane d'Ascoli, Emmanuel Chemla, Yair Lakretz, and Jean-R{\'e}mi King. 2024.
\newblock A polar coordinate system represents syntax in large language models.
\newblock \emph{Advances in Neural Information Processing Systems}, 37:105375--105396.

\bibitem[{Diessel(2023)}]{diessel2023constructicon}
Holger Diessel. 2023.
\newblock \emph{The constructicon: Taxonomies and networks}.
\newblock Cambridge University Press.

\bibitem[{Elhage et~al.(2022)Elhage, Hume, Olsson, Schiefer, Henighan, Kravec, Hatfield-Dodds, Lasenby, Drain, Chen, Grosse, McCandlish, Kaplan, Amodei, Wattenberg, and Olah}]{Elhage2022-is}
Nelson Elhage, Tristan Hume, Catherine Olsson, Nicholas Schiefer, Tom Henighan, Shauna Kravec, Zac Hatfield-Dodds, Robert Lasenby, Dawn Drain, Carol Chen, Roger Grosse, Sam McCandlish, Jared Kaplan, Dario Amodei, Martin Wattenberg, and Christopher Olah. 2022.
\newblock Toy models of superposition.
\newblock \url{https://transformer-circuits.pub/2022/toy\_model/index.html}.

\bibitem[{Elhage et~al.(2021)Elhage, Nanda, Olsson, Henighan, Joseph, Mann, Askell, Bai, Chen, Conerly, DasSarma, Drain, Ganguli, Hatfield-Dodds, Hernandez, Jones, Kernion, Lovitt, Ndousse, Amodei, Brown, Clark, Kaplan, McCandlish, and Olah}]{Elhage2021-qh}
Nelson Elhage, Neel Nanda, Catherine Olsson, Tom Henighan, Nicholas Joseph, Ben Mann, Amanda Askell, Yuntao Bai, Anna Chen, Tom Conerly, Nova DasSarma, Dawn Drain, Deep Ganguli, Zac Hatfield-Dodds, Danny Hernandez, Andy Jones, Jackson Kernion, Liane Lovitt, Kamal Ndousse, and 6 others. 2021.
\newblock A mathematical framework for transformer circuits.
\newblock \url{https://transformer-circuits.pub/2021/framework/index.html}.
\newblock Accessed: 2026-3-4.

\bibitem[{Elman(1990)}]{elman1990finding}
Jeffrey~L Elman. 1990.
\newblock Finding structure in time.
\newblock \emph{Cognitive science}, 14(2):179--211.

\bibitem[{Engels et~al.(2025)Engels, Michaud, Liao, Gurnee, and Tegmark}]{engels2025not}
Joshua Engels, Eric~J Michaud, Isaac Liao, Wes Gurnee, and Max Tegmark. 2025.
\newblock \href {https://openreview.net/forum?id=d63a4AM4hb} {Not all language model features are one-dimensionally linear}.
\newblock In \emph{The Thirteenth International Conference on Learning Representations}.

\bibitem[{Ethayarajh(2019)}]{ethayarajh-2019-contextual}
Kawin Ethayarajh. 2019.
\newblock \href {https://doi.org/10.18653/v1/D19-1006} {How contextual are contextualized word representations? {C}omparing the geometry of {BERT}, {ELM}o, and {GPT}-2 embeddings}.
\newblock In \emph{Proceedings of the 2019 Conference on Empirical Methods in Natural Language Processing and the 9th International Joint Conference on Natural Language Processing (EMNLP-IJCNLP)}, pages 55--65, Hong Kong, China. Association for Computational Linguistics.

\bibitem[{Frank(2023)}]{frank2023bridging}
Michael~C Frank. 2023.
\newblock Bridging the data gap between children and large language models.
\newblock \emph{Trends in Cognitive Sciences}.

\bibitem[{Futrell and Mahowald(2025)}]{futrell_mahowald_2025_linguists}
Richard Futrell and Kyle Mahowald. 2025.
\newblock \href {https://doi.org/10.1017/S0140525X2510112X} {How linguistics learned to stop worrying and love the language models}.
\newblock \emph{Behavioral and Brain Sciences}, page 1–98.

\bibitem[{Gauthier et~al.(2020)Gauthier, Hu, Wilcox, Qian, and Levy}]{gauthier-etal-2020-syntaxgym}
Jon Gauthier, Jennifer Hu, Ethan Wilcox, Peng Qian, and Roger Levy. 2020.
\newblock \href {https://doi.org/10.18653/v1/2020.acl-demos.10} {{S}yntax{G}ym: An online platform for targeted evaluation of language models}.
\newblock In \emph{Proceedings of the 58th Annual Meeting of the Association for Computational Linguistics: System Demonstrations}, pages 70--76, Online. Association for Computational Linguistics.

\bibitem[{Geiger et~al.(2025)Geiger, Ibeling, Zur, Chaudhary, Chauhan, Huang, Arora, Wu, Goodman, Potts et~al.}]{geiger2025causal}
Atticus Geiger, Duligur Ibeling, Amir Zur, Maheep Chaudhary, Sonakshi Chauhan, Jing Huang, Aryaman Arora, Zhengxuan Wu, Noah Goodman, Christopher Potts, and 1 others. 2025.
\newblock Causal abstraction: A theoretical foundation for mechanistic interpretability.
\newblock \emph{Journal of Machine Learning Research}, 26(83):1--64.

\bibitem[{Geiger et~al.(2022)Geiger, Wu, Lu, Rozner, Kreiss, Icard, Goodman, and Potts}]{geiger2022inducing}
Atticus Geiger, Zhengxuan Wu, Hanson Lu, Josh Rozner, Elisa Kreiss, Thomas Icard, Noah Goodman, and Christopher Potts. 2022.
\newblock Inducing causal structure for interpretable neural networks.
\newblock In \emph{International Conference on Machine Learning}, pages 7324--7338. PMLR.

\bibitem[{Geiger et~al.(2024)Geiger, Wu, Potts, Icard, and Goodman}]{geiger2024finding}
Atticus Geiger, Zhengxuan Wu, Christopher Potts, Thomas Icard, and Noah Goodman. 2024.
\newblock Finding alignments between interpretable causal variables and distributed neural representations.
\newblock In \emph{Causal Learning and Reasoning}, pages 160--187. PMLR.

\bibitem[{Gladkova et~al.(2016)Gladkova, Drozd, and Matsuoka}]{gladkova-etal-2016-analogy}
Anna Gladkova, Aleksandr Drozd, and Satoshi Matsuoka. 2016.
\newblock \href {https://doi.org/10.18653/v1/N16-2002} {Analogy-based detection of morphological and semantic relations with word embeddings: what works and what doesn{'}t.}
\newblock In \emph{Proceedings of the {NAACL} Student Research Workshop}, pages 8--15, San Diego, California. Association for Computational Linguistics.

\bibitem[{Gold(1967)}]{gold1967language}
E~Mark Gold. 1967.
\newblock Language identification in the limit.
\newblock \emph{Information and control}, 10(5):447--474.

\bibitem[{Goldberg(1995)}]{goldberg1995}
Adele~E Goldberg. 1995.
\newblock Constructions: A construction grammar approach to argument structure.
\newblock \emph{Chicago UP}.

\bibitem[{Goldberg(2003)}]{goldberg2003constructions}
Adele~E Goldberg. 2003.
\newblock Constructions: A new theoretical approach to language.
\newblock \emph{Trends in cognitive sciences}, 7(5):219--224.

\bibitem[{Grant et~al.(2026)Grant, Han, Tartaglini, and Potts}]{grant2025divergent}
Satchel Grant, Simon~Jerome Han, Alexa~R. Tartaglini, and Christopher Potts. 2026.
\newblock \href {https://openreview.net/forum?id=cZrTMqYVL6} {Addressing divergent representations from causal interventions on neural networks}.
\newblock In \emph{The Fourteenth International Conference on Learning Representations}.

\bibitem[{Hewitt and Liang(2019)}]{hewitt-liang-2019-designing}
John Hewitt and Percy Liang. 2019.
\newblock \href {https://doi.org/10.18653/v1/D19-1275} {Designing and interpreting probes with control tasks}.
\newblock In \emph{Proceedings of the 2019 Conference on Empirical Methods in Natural Language Processing and the 9th International Joint Conference on Natural Language Processing (EMNLP-IJCNLP)}, pages 2733--2743, Hong Kong, China. Association for Computational Linguistics.

\bibitem[{Hewitt and Manning(2019)}]{hewitt-manning-2019-structural}
John Hewitt and Christopher~D. Manning. 2019.
\newblock \href {https://doi.org/10.18653/v1/N19-1419} {{A} structural probe for finding syntax in word representations}.
\newblock In \emph{Proceedings of the 2019 Conference of the North {A}merican Chapter of the Association for Computational Linguistics: Human Language Technologies, Volume 1 (Long and Short Papers)}, pages 4129--4138, Minneapolis, Minnesota. Association for Computational Linguistics.

\bibitem[{Hinton(1986)}]{hinton1986learning}
Geoffrey~E Hinton. 1986.
\newblock Learning distributed representations of concepts.
\newblock In \emph{Proceedings of the Annual Meeting of the Cognitive Science Society}, volume~8.

\bibitem[{Houlsby et~al.(2019)Houlsby, Giurgiu, Jastrzebski, Morrone, De~Laroussilhe, Gesmundo, Attariyan, and Gelly}]{houlsby-peft}
Neil Houlsby, Andrei Giurgiu, Stanislaw Jastrzebski, Bruna Morrone, Quentin De~Laroussilhe, Andrea Gesmundo, Mona Attariyan, and Sylvain Gelly. 2019.
\newblock \href {https://proceedings.mlr.press/v97/houlsby19a.html} {Parameter-efficient transfer learning for {NLP}}.
\newblock In \emph{Proceedings of the 36th International Conference on Machine Learning}, volume~97 of \emph{Proceedings of Machine Learning Research}, pages 2790--2799. PMLR.

\bibitem[{Hu et~al.(2022)Hu, yelong shen, Wallis, Allen-Zhu, Li, Wang, Wang, and Chen}]{hu2022lora}
Edward~J Hu, yelong shen, Phillip Wallis, Zeyuan Allen-Zhu, Yuanzhi Li, Shean Wang, Lu~Wang, and Weizhu Chen. 2022.
\newblock \href {https://openreview.net/forum?id=nZeVKeeFYf9} {Lo{RA}: Low-rank adaptation of large language models}.
\newblock In \emph{International Conference on Learning Representations}.

\bibitem[{Hu et~al.(2024{\natexlab{a}})Hu, Mahowald, Lupyan, Ivanova, and Levy}]{hu2024}
Jennifer Hu, Kyle Mahowald, Gary Lupyan, Anna Ivanova, and Roger Levy. 2024{\natexlab{a}}.
\newblock \href {https://doi.org/10.1073/pnas.2400917121} {Language models align with human judgments on key grammatical constructions}.
\newblock \emph{Proceedings of the National Academy of Sciences}, 121(36):e2400917121.

\bibitem[{Hu et~al.(2024{\natexlab{b}})Hu, Mueller, Ross, Williams, Linzen, Zhuang, Cotterell, Choshen, Warstadt, and Wilcox}]{hu-etal-2024-findings}
Michael~Y. Hu, Aaron Mueller, Candace Ross, Adina Williams, Tal Linzen, Chengxu Zhuang, Ryan Cotterell, Leshem Choshen, Alex Warstadt, and Ethan~Gotlieb Wilcox. 2024{\natexlab{b}}.
\newblock \href {https://aclanthology.org/2024.conll-babylm.1/} {Findings of the second {B}aby{LM} challenge: Sample-efficient pretraining on developmentally plausible corpora}.
\newblock In \emph{The 2nd BabyLM Challenge at the 28th Conference on Computational Natural Language Learning}, pages 1--21, Miami, FL, USA. Association for Computational Linguistics.

\bibitem[{Hupkes et~al.(2023)Hupkes, Giulianelli, Dankers, Artetxe, Elazar, Pimentel, Christodoulopoulos, Lasri, Saphra, Sinclair et~al.}]{hupkes2023taxonomy}
Dieuwke Hupkes, Mario Giulianelli, Verna Dankers, Mikel Artetxe, Yanai Elazar, Tiago Pimentel, Christos Christodoulopoulos, Karim Lasri, Naomi Saphra, Arabella Sinclair, and 1 others. 2023.
\newblock A taxonomy and review of generalization research in nlp.
\newblock \emph{Nature Machine Intelligence}, 5(10):1161--1174.

\bibitem[{Jabbar et~al.(2025)Jabbar, Condoravdi, and Potts}]{jabbar-etal-2025-distinguishing}
Ahmad Jabbar, Cleo Condoravdi, and Christopher Potts. 2025.
\newblock \href {https://doi.org/10.18653/v1/2025.findings-emnlp.1133} {Distinguishing fair from unfair compositional generalization tasks}.
\newblock In \emph{Findings of the Association for Computational Linguistics: EMNLP 2025}, pages 20796--20807, Suzhou, China. Association for Computational Linguistics.

\bibitem[{Jackendoff(2017)}]{jackendoff2017defense}
Ray Jackendoff. 2017.
\newblock In defense of theory.
\newblock \emph{Cognitive science}, 41:185--212.

\bibitem[{Jacovi and Goldberg(2020)}]{jacovi-goldberg-2020-towards}
Alon Jacovi and Yoav Goldberg. 2020.
\newblock \href {https://doi.org/10.18653/v1/2020.acl-main.386} {Towards faithfully interpretable {NLP} systems: How should we define and evaluate faithfulness?}
\newblock In \emph{Proceedings of the 58th Annual Meeting of the Association for Computational Linguistics}, pages 4198--4205, Online. Association for Computational Linguistics.

\bibitem[{Kauf and Ivanova(2023)}]{kauf-ivanova-2023-better}
Carina Kauf and Anna~A. Ivanova. 2023.
\newblock \href {https://doi.org/10.18653/v1/2023.acl-short.80} {A better way to do masked language model scoring}.
\newblock In \emph{Proceedings of the 61st Annual Meeting of the Association for Computational Linguistics (Volume 2: Short Papers)}, pages 925--935, Toronto, Canada. Association for Computational Linguistics.

\bibitem[{Kim and Smolensky(2021)}]{kim-smolensky-2021-testing}
Najoung Kim and Paul Smolensky. 2021.
\newblock \href {https://aclanthology.org/2021.scil-1.59/} {Testing for grammatical category abstraction in neural language models}.
\newblock In \emph{Proceedings of the Society for Computation in Linguistics 2021}, pages 467--470, Online. Association for Computational Linguistics.

\bibitem[{Kingma and Ba(2014)}]{kingma2014adam}
Diederik~P Kingma and Jimmy Ba. 2014.
\newblock Adam: A method for stochastic optimization.
\newblock \emph{arXiv preprint arXiv:1412.6980}.

\bibitem[{Kumon and Yanaka(2026)}]{kumon-yanaka-2026-fine}
Ryoma Kumon and Hitomi Yanaka. 2026.
\newblock \href {https://doi.org/10.18653/v1/2026.acl-long.2078} {Fine-grained analysis of shared syntactic mechanisms in language models}.
\newblock In \emph{Proceedings of the 64th Annual Meeting of the {A}ssociation for {C}omputational {L}inguistics (Volume 1: Long Papers)}, pages 44866--44891, San Diego, California, United States. Association for Computational Linguistics.

\bibitem[{Langacker(2005)}]{langacker2005construction}
Ronald~W Langacker. 2005.
\newblock Construction grammars: Cognitive, radical, and less so.
\newblock \emph{Cognitive linguistics: Internal dynamics and interdisciplinary interaction}, 32:101--159.

\bibitem[{Leong and Linzen(2026)}]{leong-linzen-passive}
Cara Su-Yi Leong and Tal Linzen. 2026.
\newblock \href {https://doi.org/10.1016/j.jml.2026.104751} {Manipulating language models’ training data to study syntactic constraint learning: The case of english passivization}.
\newblock \emph{Journal of Memory and Language}, 149:104751.

\bibitem[{Levine and Hukari(2006)}]{Levine2006-bq}
R~Levine and T~E Hukari. 2006.
\newblock \emph{The unity of unbounded dependency constructions}.
\newblock Center for the Study of Language and Information, Stanford.

\bibitem[{Linzen et~al.(2016)Linzen, Dupoux, and Goldberg}]{linzen-etal-2016-assessing}
Tal Linzen, Emmanuel Dupoux, and Yoav Goldberg. 2016.
\newblock \href {https://doi.org/10.1162/tacl_a_00115} {Assessing the ability of {LSTM}s to learn syntax-sensitive dependencies}.
\newblock \emph{Transactions of the Association for Computational Linguistics}, 4:521--535.

\bibitem[{Liu et~al.(2019)Liu, Ott, Goyal, Du, Joshi, Chen, Levy, Lewis, Zettlemoyer, and Stoyanov}]{liu2019roberta}
Yinhan Liu, Myle Ott, Naman Goyal, Jingfei Du, Mandar Joshi, Danqi Chen, Omer Levy, Mike Lewis, Luke Zettlemoyer, and Veselin Stoyanov. 2019.
\newblock \href {https://arxiv.org/abs/1907.11692} {{RoBERTa}: A robustly optimized bert pretraining approach}.
\newblock \emph{Preprint}, arXiv:1907.11692.

\bibitem[{Loureiro et~al.(2021)Loureiro, Rezaee, Pilehvar, and Camacho-Collados}]{loureiro-etal-2021-analysis}
Daniel Loureiro, Kiamehr Rezaee, Mohammad~Taher Pilehvar, and Jose Camacho-Collados. 2021.
\newblock \href {https://doi.org/10.1162/coli_a_00405} {Analysis and evaluation of language models for word sense disambiguation}.
\newblock \emph{Computational Linguistics}, 47(2):387--443.

\bibitem[{Makelov et~al.(2024)Makelov, Lange, Geiger, and Nanda}]{makelov2024illusion}
Aleksandar Makelov, Georg Lange, Atticus Geiger, and Neel Nanda. 2024.
\newblock \href {https://openreview.net/forum?id=Ebt7JgMHv1} {Is this the subspace you are looking for? an interpretability illusion for subspace activation patching}.
\newblock In \emph{The Twelfth International Conference on Learning Representations}.

\bibitem[{Marantz and Halle(1993)}]{Marantz1993-wv}
Alec Marantz and Morris Halle. 1993.
\newblock Distributed morphology and the pieces of inflection.
\newblock In Kenneth~Locke Hale and Samuel~Jay Keyser, editors, \emph{The View from Building 20: Essays in Linguistics in Honor of Sylvain Bromberger}, pages 112--176. MIT Press, Cambridge.

\bibitem[{Marks and Tegmark(2024)}]{marks2024means}
Samuel Marks and Max Tegmark. 2024.
\newblock \href {https://openreview.net/forum?id=aajyHYjjsk} {The geometry of truth: Emergent linear structure in large language model representations of true/false datasets}.
\newblock In \emph{First Conference on Language Modeling}.

\bibitem[{McCoy et~al.(2019)McCoy, Linzen, Dunbar, and Smolensky}]{mccoy2018rnns}
R.~Thomas McCoy, Tal Linzen, Ewan Dunbar, and Paul Smolensky. 2019.
\newblock \href {https://openreview.net/forum?id=BJx0sjC5FX} {{RNN}s implicitly implement tensor-product representations}.
\newblock In \emph{International Conference on Learning Representations}.

\bibitem[{M{\'e}loux et~al.(2025)M{\'e}loux, Maniu, Portet, and Peyrard}]{meloux2025everything}
Maxime M{\'e}loux, Silviu Maniu, Fran{\c{c}}ois Portet, and Maxime Peyrard. 2025.
\newblock \href {https://openreview.net/forum?id=5IWJBStfU7} {Everything, everywhere, all at once: Is mechanistic interpretability identifiable?}
\newblock In \emph{The Thirteenth International Conference on Learning Representations}.

\bibitem[{Merity et~al.(2017)Merity, Xiong, Bradbury, and Socher}]{merity2017pointer}
Stephen Merity, Caiming Xiong, James Bradbury, and Richard Socher. 2017.
\newblock \href {https://openreview.net/forum?id=Byj72udxe} {Pointer sentinel mixture models}.
\newblock In \emph{International Conference on Learning Representations}.

\bibitem[{Mikolov et~al.(2013)Mikolov, Sutskever, Chen, Corrado, and Dean}]{mikolov2013b}
Tom{\'a}{\v{s}} Mikolov, Ilya Sutskever, Kai Chen, Greg~S Corrado, and Jeff Dean. 2013.
\newblock Distributed representations of words and phrases and their compositionality.
\newblock \emph{Advances in neural information processing systems}, 26.

\bibitem[{Millière(2024)}]{milliere2024LMs}
Raphaël Millière. 2024.
\newblock \href {https://arxiv.org/abs/2408.07144} {Language models as models of language}.
\newblock \emph{Preprint}, arXiv:2408.07144.

\bibitem[{Misra(2022)}]{misra2022minicons}
Kanishka Misra. 2022.
\newblock minicons: Enabling flexible behavioral and representational analyses of transformer language models.
\newblock \emph{arXiv preprint arXiv:2203.13112}.

\bibitem[{Misra and Kim(2023)}]{misra2023abstraction}
Kanishka Misra and Najoung Kim. 2023.
\newblock \href {https://arxiv.org/abs/2312.03708} {Abstraction via exemplars? a representational case study on lexical category inference in bert}.
\newblock \emph{Preprint}, arXiv:2312.03708.

\bibitem[{Misra and Mahowald(2024)}]{misra-mahowald-2024-language}
Kanishka Misra and Kyle Mahowald. 2024.
\newblock \href {https://doi.org/10.18653/v1/2024.emnlp-main.53} {Language models learn rare phenomena from less rare phenomena: The case of the missing {AANN}s}.
\newblock In \emph{Proceedings of the 2024 Conference on Empirical Methods in Natural Language Processing}, pages 913--929, Miami, Florida, USA. Association for Computational Linguistics.

\bibitem[{Mueller et~al.(2026)Mueller, Brinkmann, Li, Marks, Pal, Prakash, Rager, Sankaranarayanan, Sen~Sharma, Sun, Todd, Bau, and Belinkov}]{mueller-etal-2026-quest}
Aaron Mueller, Jannik Brinkmann, Millicent Li, Samuel Marks, Koyena Pal, Nikhil Prakash, Can Rager, Aruna Sankaranarayanan, Arnab Sen~Sharma, Jiuding Sun, Eric Todd, David Bau, and Yonatan Belinkov. 2026.
\newblock \href {https://doi.org/10.1162/coli.a.572} {The quest for the right mediator: Surveying mechanistic interpretability for {NLP} through the lens of causal mediation analysis}.
\newblock \emph{Computational Linguistics}, 52(1):331--378.

\bibitem[{Nanda et~al.(2023)Nanda, Lee, and Wattenberg}]{nanda-etal-2023-emergent}
Neel Nanda, Andrew Lee, and Martin Wattenberg. 2023.
\newblock \href {https://doi.org/10.18653/v1/2023.blackboxnlp-1.2} {Emergent linear representations in world models of self-supervised sequence models}.
\newblock In \emph{Proceedings of the 6th BlackboxNLP Workshop: Analyzing and Interpreting Neural Networks for NLP}, pages 16--30, Singapore. Association for Computational Linguistics.

\bibitem[{Pacton and Peereman(2023)}]{pacton2023morphology}
S{\'e}bastien Pacton and Ronald Peereman. 2023.
\newblock Morphology as an aid in orthographic learning of new words: The influence of inflected and derived forms in spelling acquisition.
\newblock \emph{Journal of Experimental Child Psychology}, 232:105675.

\bibitem[{Park et~al.(2024)Park, Choe, and Veitch}]{park2024linear}
Kiho Park, Yo~Joong Choe, and Victor Veitch. 2024.
\newblock The linear representation hypothesis and the geometry of large language models.
\newblock In \emph{Proceedings of the 41st International Conference on Machine Learning}, pages 39643--39666.

\bibitem[{Patil et~al.(2024)Patil, Jumelet, Chiu, Lapastora, Shen, Wang, Willrich, and Steinert-Threlkeld}]{patil-etal-2024-filtered}
Abhinav Patil, Jaap Jumelet, Yu~Ying Chiu, Andy Lapastora, Peter Shen, Lexie Wang, Clevis Willrich, and Shane Steinert-Threlkeld. 2024.
\newblock \href {https://doi.org/10.1162/tacl_a_00720} {Filtered corpus training ({F}i{CT}) shows that language models can generalize from indirect evidence}.
\newblock \emph{Transactions of the Association for Computational Linguistics}, 12:1597--1615.

\bibitem[{Pedregosa et~al.(2011)Pedregosa, Varoquaux, Gramfort, Michel, Thirion, Grisel, Blondel, Prettenhofer, Weiss, Dubourg, Vanderplas, Passos, Cournapeau, Brucher, Perrot, and Duchesnay}]{scikit-learn}
F.~Pedregosa, G.~Varoquaux, A.~Gramfort, V.~Michel, B.~Thirion, O.~Grisel, M.~Blondel, P.~Prettenhofer, R.~Weiss, V.~Dubourg, J.~Vanderplas, A.~Passos, D.~Cournapeau, M.~Brucher, M.~Perrot, and E.~Duchesnay. 2011.
\newblock Scikit-learn: Machine learning in {P}ython.
\newblock \emph{Journal of Machine Learning Research}, 12:2825--2830.

\bibitem[{Pimentel et~al.(2020)Pimentel, Valvoda, Maudslay, Zmigrod, Williams, and Cotterell}]{pimentel-etal-2020-information}
Tiago Pimentel, Josef Valvoda, Rowan~Hall Maudslay, Ran Zmigrod, Adina Williams, and Ryan Cotterell. 2020.
\newblock \href {https://doi.org/10.18653/v1/2020.acl-main.420} {Information-theoretic probing for linguistic structure}.
\newblock In \emph{Proceedings of the 58th Annual Meeting of the Association for Computational Linguistics}, pages 4609--4622, Online. Association for Computational Linguistics.

\bibitem[{Reif et~al.(2019)Reif, Yuan, Wattenberg, Viegas, Coenen, Pearce, and Kim}]{reif2019visualizing}
Emily Reif, Ann Yuan, Martin Wattenberg, Fernanda~B Viegas, Andy Coenen, Adam Pearce, and Been Kim. 2019.
\newblock Visualizing and measuring the geometry of bert.
\newblock \emph{Advances in neural information processing systems}, 32.

\bibitem[{Rogers et~al.(2020)Rogers, Kovaleva, and Rumshisky}]{rogers-etal-2020-primer}
Anna Rogers, Olga Kovaleva, and Anna Rumshisky. 2020.
\newblock \href {https://doi.org/10.1162/tacl_a_00349} {A primer in {BERT}ology: What we know about how {BERT} works}.
\newblock \emph{Transactions of the Association for Computational Linguistics}, 8:842--866.

\bibitem[{Ross(1967)}]{Ross1967-je}
John~Robert Ross. 1967.
\newblock \emph{Constraints on variables in syntax}.
\newblock Ph.D. thesis, MIT.

\bibitem[{Rozner et~al.(2021)Rozner, Potts, and Mahowald}]{roznercryptics}
Josh Rozner, Christopher Potts, and Kyle Mahowald. 2021.
\newblock \href {https://proceedings.neurips.cc/paper_files/paper/2021/file/5f1d3986fae10ed2994d14ecd89892d7-Paper.pdf} {Decrypting cryptic crosswords: Semantically complex wordplay puzzles as a target for nlp}.
\newblock In \emph{Advances in Neural Information Processing Systems}, volume~34, pages 11409--11421. Curran Associates, Inc.

\bibitem[{Rozner et~al.(2025{\natexlab{a}})Rozner, Weissweiler, Mahowald, and Shain}]{rozner-etal-2025-constructions}
Joshua Rozner, Leonie Weissweiler, Kyle Mahowald, and Cory Shain. 2025{\natexlab{a}}.
\newblock \href {https://doi.org/10.18653/v1/2025.emnlp-main.108} {Constructions are revealed in word distributions}.
\newblock In \emph{Proceedings of the 2025 Conference on Empirical Methods in Natural Language Processing}, pages 2116--2138, Suzhou, China. Association for Computational Linguistics.

\bibitem[{Rozner et~al.(2025{\natexlab{b}})Rozner, Weissweiler, and Shain}]{rozner-etal-2025-babylms}
Joshua Rozner, Leonie Weissweiler, and Cory Shain. 2025{\natexlab{b}}.
\newblock \href {https://doi.org/10.18653/v1/2025.emnlp-main.113} {{B}aby{LM}{'}s first constructions: Causal interventions provide a signal of learning}.
\newblock In \emph{Proceedings of the 2025 Conference on Empirical Methods in Natural Language Processing}, pages 2237--2249, Suzhou, China. Association for Computational Linguistics.

\bibitem[{Sag(2010)}]{sag2010english}
Ivan~A Sag. 2010.
\newblock English filler-gap constructions.
\newblock \emph{Language}, pages 486--545.

\bibitem[{Saphra and Wiegreffe(2024)}]{saphra-wiegreffe-2024-mechanistic}
Naomi Saphra and Sarah Wiegreffe. 2024.
\newblock \href {https://doi.org/10.18653/v1/2024.blackboxnlp-1.30} {Mechanistic?}
\newblock In \emph{Proceedings of the 7th BlackboxNLP Workshop: Analyzing and Interpreting Neural Networks for NLP}, pages 480--498, Miami, Florida, US. Association for Computational Linguistics.

\bibitem[{Shadish et~al.(2002)Shadish, Cook, and Campbell}]{shadish2002experimental}
William~R. Shadish, Thomas~D. Cook, and Donald~T. Campbell. 2002.
\newblock \emph{Experimental and Quasi-Experimental Designs for Generalized Causal Inference}.
\newblock Houghton Mifflin Company, Boston, MA.

\bibitem[{Shepard(1987)}]{shepard1987toward}
Roger~N Shepard. 1987.
\newblock Toward a universal law of generalization for psychological science.
\newblock \emph{Science}, 237(4820):1317--1323.

\bibitem[{Smolensky(1990)}]{smolensky_tensor_1990}
Paul Smolensky. 1990.
\newblock Tensor product variable binding and the representation of symbolic structures in connectionist systems.
\newblock \emph{Artificial intelligence}, 46(1-2):159--216.
\newblock Publisher: Elsevier.

\bibitem[{Springer et~al.(2025)Springer, Kotha, Fried, Neubig, and Raghunathan}]{springer2025repetition}
Jacob Springer, Suhas Kotha, Daniel Fried, Graham Neubig, and Aditi Raghunathan. 2025.
\newblock Repetition improves language model embeddings.
\newblock In \emph{International Conference on Learning Representations}, volume 2025, pages 93543--93579.

\bibitem[{Stanczak et~al.(2022)Stanczak, Ponti, Torroba~Hennigen, Cotterell, and Augenstein}]{stanczak-etal-2022-neurons}
Karolina Stanczak, Edoardo Ponti, Lucas Torroba~Hennigen, Ryan Cotterell, and Isabelle Augenstein. 2022.
\newblock \href {https://doi.org/10.18653/v1/2022.naacl-main.114} {Same neurons, different languages: Probing morphosyntax in multilingual pre-trained models}.
\newblock In \emph{Proceedings of the 2022 Conference of the North American Chapter of the Association for Computational Linguistics: Human Language Technologies}, pages 1589--1598, Seattle, United States. Association for Computational Linguistics.

\bibitem[{Steck et~al.(2024)Steck, Ekanadham, and Kallus}]{steck2024cosine}
Harald Steck, Chaitanya Ekanadham, and Nathan Kallus. 2024.
\newblock Is cosine-similarity of embeddings really about similarity?
\newblock In \emph{Companion Proceedings of the ACM Web Conference 2024}, pages 887--890.

\bibitem[{Steele et~al.(2026)Steele, Wen, and Han}]{steele2026das}
Oliver Steele, Jiangtao Wen, and Yuxing Han. 2026.
\newblock \href {https://arxiv.org/abs/2607.10248} {One mechanism for many mental spaces: a shared router over a value slot in language models}.
\newblock \emph{Preprint}, arXiv:2607.10248.

\bibitem[{Stefanowitsch and Gries(2003)}]{stefanowitsch2003}
Anatol Stefanowitsch and Stefan~Th. Gries. 2003.
\newblock \href {https://doi.org/10.1075/ijcl.8.2.03ste} {Collostructions: Investigating the interaction of words and constructions}.
\newblock \emph{International Journal of Corpus Linguistics}, 8(2):209--243.

\bibitem[{Sutter et~al.(2025)Sutter, Minder, Hofmann, and Pimentel}]{sutter2025non}
Denis Sutter, Julian Minder, Thomas Hofmann, and Tiago Pimentel. 2025.
\newblock \href {https://openreview.net/forum?id=ZYXTLo7kCi} {The non-linear representation dilemma: Is causal abstraction enough for mechanistic interpretability?}
\newblock In \emph{The Thirty-ninth Annual Conference on Neural Information Processing Systems}.

\bibitem[{Teglia et~al.(2025)Teglia, Tedeschi, and Navigli}]{teglia-etal-2025-much}
Simone Teglia, Simone Tedeschi, and Roberto Navigli. 2025.
\newblock \href {https://doi.org/10.18653/v1/2025.acl-long.113} {How much do encoder models know about word senses?}
\newblock In \emph{Proceedings of the 63rd Annual Meeting of the Association for Computational Linguistics (Volume 1: Long Papers)}, pages 2266--2277, Vienna, Austria. Association for Computational Linguistics.

\bibitem[{Templeton et~al.(2024)Templeton, Conerly, Marcus, Lindsey, Bricken, Chen, Pearce, Citro, Ameisen, Jones, Cunningham, Turner, McDougall, MacDiarmid, Tamkin, Durmus, Hume, Mosconi, Daniel~Freeman, Sumers, Rees, Batson, Jermyn, Carter, Olah, and Henighan}]{Templeton2024-hs}
Adly Templeton, Tom Conerly, Jonathan Marcus, Jack Lindsey, Trenton Bricken, Brian Chen, Adam Pearce, Craig Citro, Emmanuel Ameisen, Andy Jones, Hoagy Cunningham, Nicholas~L Turner, Callum McDougall, Monte MacDiarmid, Alex Tamkin, Esin Durmus, Tristan Hume, Francesco Mosconi, C~Daniel~Freeman, and 7 others. 2024.
\newblock Scaling monosemanticity: Extracting interpretable features from claude 3 sonnet.
\newblock \url{https://transformer-circuits.pub/2024/scaling-monosemanticity/}.
\newblock Accessed: 2026-3-4.

\bibitem[{Tenenbaum et~al.(2011)Tenenbaum, Kemp, Griffiths, and Goodman}]{Tenenbaum2011-ev}
Joshua~B Tenenbaum, Charles Kemp, Thomas~L Griffiths, and Noah~D Goodman. 2011.
\newblock How to grow a mind: statistics, structure, and abstraction.
\newblock \emph{Science}, 331(6022):1279--1285.

\bibitem[{Tenney et~al.(2019)Tenney, Das, and Pavlick}]{tenney-etal-2019-bert}
Ian Tenney, Dipanjan Das, and Ellie Pavlick. 2019.
\newblock \href {https://doi.org/10.18653/v1/P19-1452} {{BERT} rediscovers the classical {NLP} pipeline}.
\newblock In \emph{Proceedings of the 57th Annual Meeting of the Association for Computational Linguistics}, pages 4593--4601, Florence, Italy. Association for Computational Linguistics.

\bibitem[{Timkey and van Schijndel(2021)}]{timkey-van-schijndel-2021-bark}
William Timkey and Marten van Schijndel. 2021.
\newblock \href {https://doi.org/10.18653/v1/2021.emnlp-main.372} {All bark and no bite: Rogue dimensions in transformer language models obscure representational quality}.
\newblock In \emph{Proceedings of the 2021 Conference on Empirical Methods in Natural Language Processing}, pages 4527--4546, Online and Punta Cana, Dominican Republic. Association for Computational Linguistics.

\bibitem[{Tomasello(2005)}]{tomasello2005constructing}
Michael Tomasello. 2005.
\newblock \emph{Constructing a language: A usage-based theory of language acquisition}.
\newblock Harvard university press.

\bibitem[{Trinley et~al.(2026)Trinley, Alabi, Klakow, and Gautam}]{trinley2026das}
Katharina Trinley, Jesujoba~O. Alabi, Dietrich Klakow, and Vagrant Gautam. 2026.
\newblock \href {https://arxiv.org/abs/2606.16407} {A mechanistic understanding of pronoun fidelity in llms}.
\newblock \emph{Preprint}, arXiv:2606.16407.

\bibitem[{Tucker et~al.(2021)Tucker, Qian, and Levy}]{tucker-etal-2021-modified}
Mycal Tucker, Peng Qian, and Roger Levy. 2021.
\newblock \href {https://doi.org/10.18653/v1/2021.findings-acl.76} {What if this modified that? syntactic interventions with counterfactual embeddings}.
\newblock In \emph{Findings of the Association for Computational Linguistics: ACL-IJCNLP 2021}, pages 862--875, Online. Association for Computational Linguistics.

\bibitem[{Warner et~al.(2025)Warner, Chaffin, Clavi{\'e}, Weller, Hallstr{\"o}m, Taghadouini, Gallagher, Biswas, Ladhak, Aarsen, Adams, Howard, and Poli}]{warner-etal-2025-smarter}
Benjamin Warner, Antoine Chaffin, Benjamin Clavi{\'e}, Orion Weller, Oskar Hallstr{\"o}m, Said Taghadouini, Alexis Gallagher, Raja Biswas, Faisal Ladhak, Tom Aarsen, Griffin~Thomas Adams, Jeremy Howard, and Iacopo Poli. 2025.
\newblock \href {https://doi.org/10.18653/v1/2025.acl-long.127} {Smarter, better, faster, longer: A modern bidirectional encoder for fast, memory efficient, and long context finetuning and inference}.
\newblock In \emph{Proceedings of the 63rd Annual Meeting of the Association for Computational Linguistics (Volume 1: Long Papers)}, pages 2526--2547, Vienna, Austria. Association for Computational Linguistics.

\bibitem[{Warstadt and Bowman(2022)}]{warstadt2022artificial}
Alex Warstadt and Samuel~R Bowman. 2022.
\newblock What artificial neural networks can tell us about human language acquisition.
\newblock In \emph{Algebraic structures in natural language}, pages 17--60. CRC Press.

\bibitem[{Warstadt et~al.(2023)Warstadt, Mueller, Choshen, Wilcox, Zhuang, Ciro, Mosquera, Paranjabe, Williams, Linzen, and Cotterell}]{warstadt-etal-2023-findings}
Alex Warstadt, Aaron Mueller, Leshem Choshen, Ethan Wilcox, Chengxu Zhuang, Juan Ciro, Rafael Mosquera, Bhargavi Paranjabe, Adina Williams, Tal Linzen, and Ryan Cotterell. 2023.
\newblock \href {https://doi.org/10.18653/v1/2023.conll-babylm.1} {Findings of the {B}aby{LM} challenge: Sample-efficient pretraining on developmentally plausible corpora}.
\newblock In \emph{Proceedings of the BabyLM Challenge at the 27th Conference on Computational Natural Language Learning}, pages 1--34, Singapore. Association for Computational Linguistics.

\bibitem[{White et~al.(2021)White, Pimentel, Saphra, and Cotterell}]{white-etal-2021-non}
Jennifer~C. White, Tiago Pimentel, Naomi Saphra, and Ryan Cotterell. 2021.
\newblock \href {https://doi.org/10.18653/v1/2021.naacl-main.12} {A non-linear structural probe}.
\newblock In \emph{Proceedings of the 2021 Conference of the North American Chapter of the Association for Computational Linguistics: Human Language Technologies}, pages 132--138, Online. Association for Computational Linguistics.

\bibitem[{Wolf et~al.(2019)Wolf, Debut, Sanh, Chaumond, Delangue, Moi, Cistac, Rault, Louf, Funtowicz, and Brew}]{huggingface}
Thomas Wolf, Lysandre Debut, Victor Sanh, Julien Chaumond, Clement Delangue, Anthony Moi, Pierric Cistac, Tim Rault, R{\'{e}}mi Louf, Morgan Funtowicz, and Jamie Brew. 2019.
\newblock \href {https://arxiv.org/abs/1910.03771} {Huggingface's transformers: State-of-the-art natural language processing}.
\newblock \emph{CoRR}, abs/1910.03771.

\bibitem[{Wu et~al.(2024)Wu, Geiger, Huang, Arora, Icard, Potts, and Goodman}]{wu2024replymakelove}
Zhengxuan Wu, Atticus Geiger, Jing Huang, Aryaman Arora, Thomas Icard, Christopher Potts, and Noah~D. Goodman. 2024.
\newblock \href {https://arxiv.org/abs/2401.12631} {A reply to makelov et al. (2023)'s "interpretability illusion" arguments}.
\newblock \emph{Preprint}, arXiv:2401.12631.

\bibitem[{Wu et~al.(2023)Wu, Geiger, Icard, Potts, and Goodman}]{wu2023interpretability}
Zhengxuan Wu, Atticus Geiger, Thomas Icard, Christopher Potts, and Noah Goodman. 2023.
\newblock \href {https://openreview.net/forum?id=nRfClnMhVX} {Interpretability at scale: Identifying causal mechanisms in alpaca}.
\newblock In \emph{Thirty-seventh Conference on Neural Information Processing Systems}.

\bibitem[{Zhang and Nanda(2024)}]{zhang2024towards}
Fred Zhang and Neel Nanda. 2024.
\newblock \href {https://openreview.net/forum?id=Hf17y6u9BC} {Towards best practices of activation patching in language models: Metrics and methods}.
\newblock In \emph{The Twelfth International Conference on Learning Representations}.

\bibitem[{Zhou et~al.(2022)Zhou, Ethayarajh, Card, and Jurafsky}]{zhou-etal-2022-problems}
Kaitlyn Zhou, Kawin Ethayarajh, Dallas Card, and Dan Jurafsky. 2022.
\newblock \href {https://doi.org/10.18653/v1/2022.acl-short.45} {Problems with cosine as a measure of embedding similarity for high frequency words}.
\newblock In \emph{Proceedings of the 60th Annual Meeting of the Association for Computational Linguistics (Volume 2: Short Papers)}, pages 401--423, Dublin, Ireland. Association for Computational Linguistics.

\end{thebibliography}

\appendix
\label{sec:appendix}
\renewcommand{\thetable}{A\arabic{table}}
\setcounter{table}{0}
\renewcommand{\thefigure}{A\arabic{figure}}
\setcounter{figure}{0}

\section{Supplement: Methods}
\label{sec:app:methods}

\subsection{Models}
We use GPT-BERT as our BabyLM because  
it was the best-performing model in the 2024 BabyLM challenge \cite{hu-etal-2024-findings} and is an MLM.
We use the version of it that was used as a baseline in the 2025 BabyLM challenge \cite{charpentier-etal-2025-findings}.\footnote{
Accessed via HuggingFace: \url{https://huggingface.co/BabyLM-community/babylm-baseline-100m-gpt-bert-mixed}
}
Other models (RoBERTa, modernBERT, and Pythia) are also obtained from Huggingface.
For GPT-BERT, RoBERTa and modernBERT, we use Huggingface's \texttt{from\_config} method to obtain a randomly initialized version of the model.
For the untrained Pythia model, we use the untrained step 0 checkpoint.

We perturb (i.e. train) using \texttt{AdamW}, with all standard arguments except the learning rate; see below.

\subsection{Hyperparameter selection}
\label{sec:app:hyper}
As each model has a slightly different size and structure, we tune perturbation parameters (learning rate and number of steps) against AUC clusterability (see \sref{sec:methods:exp}, \aref{app:methods:auc}; either separating the linguistic classes of interest, or distinguishing a control class) using a separate training set before conducting final evaluation on a held-out test set.
(For Lexical Representations and Filler Gap, train-test splits were provided by the existing benchmarks.
For Morphology, we create the separate sets via separate samples of the Cosmo Corpus.)
Details of the classes and controls are provided in each experiment section and final hyperparameter choices are in each experiment section's corresponding Appendix.

Though we use separate train and test sets to make sure that we do not tune the perturbation parameters to the train set, performance across training and test sets is similar and there is no evidence that, in optimizing steps and learning rate, perturbation ``learns the train set.''
For example, for morphology, RoBERTa-large has AUC 0.898 on the training set and 0.889 on the test set. 
modernBERT-large has AUC 0.858 on training and 0.871 on test. 

In the hyperparameter search, we observe that perturbation achieves high clusterability on trained models at a variety of learning rates and gradient steps.
Unsurprisingly, high learning rates or too many steps can collapse model behavior; lower learning rates produce smaller effects.

In \sectionref{sec:fg} we reported how much model perplexity changes after perturbation on the filler-gap syntax task.
Unsurprisingly, during hyperparameter tuning on the filler-gap syntax task, we observe that higher learning rates and more gradient steps increase perplexity more than lower learning rates and fewer steps. 

\subsection{Cost of perturbation training and evaluation}
\label{sec:app:methods:cost}
Perturbation is relatively inexpensive to run.
For example, for \sectionref{sec:fg}, we can run a full experiment on a Pythia-1.4B checkpoint in 15 minutes on a single Nvidia RTX A6000. 
This produces 145k observations: 5 separate training perturbations of each of 34 classes (170 perturbations), with each evaluated on 25 examples from each class (150 data points per grid square in the similarity matrices produced in \autoref{fig:app:fg_acq_all}).

For comparison, \citet{boguraev-etal-2025-causal} report that for one model checkpoint, training and evaluation on two Nvidia A40 GPUs required 262 hours. 
Our entire FG analysis (which evaluated 13 model checkpoints on both FG and the non-FG control) took $15 \text{min} \times 13 \times 2 = $ 7hrs.

\subsection{Clusterability AUC}
\label{app:methods:auc}
As noted in \sectionref{sec:methods:exp}, we use a generic notion of \textit{clusterability} for evaluation, which measures the degree to which similarities are higher within vs. between classes as an area under the curve.
Specifically, we compute an $n \times n$ perturbation transfer  matrix (i.e., a perturbation-based similarity matrix) across all examples in the dataset, where the value at cell $i, j$ is the log odds-ratio $\mathcal{R}$ when the model is evaluated on remapping $j$ after being perturbed (i.e., trained) using remapping $i$.
We treat these scores as a classifier signal for ``same cluster'' and compute the area under the curve (AUC) of the receiver operating characteristic against a label matrix $L$ where $L_{i,j} = 1$ iff $i$ and $j$ are of the same class, and zero otherwise.\footnote{
We use scikit \cite{scikit-learn}.
Although AUC is invariant to the ratio of positive to negative labels, it is sensitive to the distribution of subclasses \emph{within} the positive and negative sets.
As the distribution of subclasses in the dataset is not always balanced, we compute AUC with reweighting, so that each subclass contributes equally within both the positive and negative label sets.
}
In this setting, a higher AUC indicates that same-class items tend to have higher similarity scores than different-class items.
Compared to analyses based on actual clustering, clusterability has the benefit that it produces a deterministic value that does not depend on the choice of a particular clustering technique or seed.

\section{Supplement for \sectionref{sec:morpho}}
\subsection{Methods}
As described in \sectionref{sec:morpho}, we evaluate on a test set of 98 examples (one each of the $2 \times 50 = 100$ -er words, except for two words which were not found in the BabyLM Cosmo corpus). 
Hyperparameters (learning rate and number of optimizer steps) were tuned on separate data for training to maximize AUC clusterability across the two classes.
Hyperparameters for all models tested are given in \aref{sec:app:morph:other_models}.

Although the linear model computed after symmetrizing the effects matrix (i.e. averaging (i,j) with (j,i)), for AUC clusterability we do not symmetrize.

\subsection{Linear model}
\label{sec:app:morph:lm}

\begin{table*}[h]
\begin{NiceTabular}{m{0.7cm}m{0.8cm} l c c}
\toprule
\textbf{trained} & \textbf{class} & \textbf{tokenization} &
\textbf{effect} &\textbf{p-val}\\
\midrule

\Block{8-1}{false}
  & \Block{4-1}{between class}
    & multi-diff   & $-0.10 \pm 0.03$ & $0.003$\\
  &   & multi-same   & \textbf{3.3}$ \pm 0.2$ & $<0.001$ \\
  &   & single-multi & $-0.06 \pm 0.02$ & $0.009$\\
  &   & single-single& $-0.09 \pm 0.05$ & $0.04$\\
\cmidrule(lr){2-5}
  & \Block{4-1}{within class}
    & multi-diff   & $-0.17\pm 0.06$ & $0.05$\\
  &   & multi-same   & \textbf{2.6} $\pm 0.2$ &$<0.001$  \\
  &   & single-multi & $-0.10 \pm 0.04$ & $0.03$\\
  &   & single-single& $-0.10 \pm 0.08$ & $0.02$\\
\midrule

\Block{8-1}{true}
  & \Block{4-1}{between class}
    & multi-diff   & $0.03 \pm 0.03$ & $0.03$\\
  &   & multi-same   & $0.04 \pm 0.2$ & $0.09$\\
  &   & single-multi & $-0.14 \pm 0.02$& $<0.001$\\
  &   & single-single& $0.08 \pm 0.05$ & $0.06$\\
\cmidrule(lr){2-5}
  & \Block{4-1}{within class}
    & multi-diff   & \textbf{0.95} $ \pm 0.06$ & $<0.001$\\
  &   & multi-same   & \textbf{1.3} $ \pm 0.2$  & $<0.001$\\
  &   & single-multi & \textbf{1.34} $\pm 0.04$ & $<0.001$\\
  &   & single-single& \textbf{3.32} $ \pm 0.08$ & $<0.001$\\
\bottomrule
\end{NiceTabular}
\caption{
Morphology Linear Model Summary: Average effects by all three factors.
Average effects greater than 0.1 are bolded.
As discussed in \sref{sec:morpho}, untrained models show meaningful transfer only in the multi-same (shared token) setting.
Trained models show meaningful transfer only to items of the same class.
In trained models single-single transfer is significantly higher than other tokenization settings (discussed in \aref{app:morpho:addnal}).
}
\label{tab:app:effect_all}
\end{table*}

\begin{table}[t]
\begin{NiceTabular}{lccc}
\toprule
 & \textbf{trained} & \textbf{untrained} & \textbf{diff} \\
\midrule
within class 
& \makecell{1.73 \\ ($p<0.001$)}
& \makecell{0.56 \\ ($p<0.001$)}
& \makecell{1.17 \\ ($p<0.001$)} \\

between class
& \makecell{0.003 \\ ($p=0.97$)}
& \makecell{0.75 \\ ($p<0.001$)}
& \makecell{-0.75 \\ ($p<0.001$)} \\
\bottomrule
\end{NiceTabular}
\caption{
Marginal effects by class and training status.
(Note: since we use balanced factor weights, these correspond to the average of the corresponding values in \autoref{tab:app:effect_all}.)
Trained models show transfer only within class.
Untrained models show some transfer both within and between classes, but this is because the average includes pairs with shared tokens (see \autoref{tab:app:morpho_tokenization}).
}
\label{tab:app:morpho_avg_effect}
\end{table}

\begin{table}[t]
\begin{NiceTabular}{llcc}
\toprule
 &  & \textbf{trained} & \textbf{untrained} \\
\midrule
\Block{2-1}{Shared token}
  & within class  &  1.3 &  2.6 \\
  & between class &  \makecell{0.04 \\($p=0.9$)} &  3.3 \\
\midrule
\Block{2-1}{No shared token}
  & within class  &  1.87 & -0.12 \\
  & between class & \makecell{-0.01 \\ ($p=0.6$)} & -0.08 \\
\bottomrule
\end{NiceTabular}
\caption{
Marginal effects by whether there is a shared token (i.e., multi-same is True).
This table can be computed from \autoref{tab:app:effect_all}, with averaging in the case of within-class (deverbal and comparative) and no-shared-token (average of the three values for the tokenization factor that are not multi-same).
Untrained models show transfer (both within and between classes) only when there is a shared token.
Trained models show similar transfer effect regardless of whether there is a shared token.
In trained models, effect size with no shared token (1.87) is higher than shared token (1.3) because the single-single tokenization setting has higher transfer than all multi-token settings.
P-values are $<0.001$ except where indicated.
}
\label{tab:app:morpho_tokenization}
\end{table}

Marginal effects are computed with the \margeff package \cite{marginaleffects} using balanced factor weights.
\autoref{tab:app:effect_all} gives average effects across the three factors tested in the linear model in \sref{sec:morpho} (though it omits for clarity the distinction between same-class-comparative and same-class-deverbal).
\autoref{tab:app:morpho_avg_effect} and \autoref{tab:app:morpho_tokenization} give average effects for the aggregations discussed in that section.

\subsection{Vertical patterning in untrained models}
\label{app:morpho:permutations}

We verify that vertical patterning in the untrained model is random.
We initialize a HuggingFace RoBERTa-large model with seven different seeds.
We compute the transfer matrix on the common training dataset for all of these models.

We look for linear dependence in the transfer matrix structure (i.e. consistent vertical patterning across initializations) using a permutation test.
We consider the first $33 \times 33$ block of each matrix (this is the portion of the matrix before any shared tokens). 
We first compute the average observed Pearson correlation coefficient (on flattened matrices with diagonal excluded) over all pairs of the 7 transfer matrices (21 unique pairs).

We test under column permutation and also full matrix permutation.
For column permutation, we construct a null distribution by computing the same statistic (i.e. average of correlation coefficient over 21 matrix pairs) under n=10,000 random permutations of the matrix columns, which results in a p-value of 0.720 (i.e., the permuted matrices produce a statistic larger than the observed statistic in 72\% of trials).

For the full matrix permutation, we do the same but randomly permute the entire matrix (we flatten, shuffle, and then unflatten).
Under n=10,000 random permutations of the matrices, we obtain a p-value of 0.725.
We thus fail to find evidence of systematic patterning in the transfer to particular items when there are no shared tokens.

\subsection{Additional findings from the morphology analysis}
\label{app:morpho:addnal}

\paragraph{Multi-token effects}
In trained models, within-class pairs show significantly higher transfer when both are single-token: single-single (3.32) vs. multi-diff (0.95; $\Delta = 2.37, p<0.001$; see \autoref{tab:app:effect_all}). 
This suggests either that our approach to perturbation does not work as well in the multi-token setting, or that shared representations are weaker in multitoken words.

\begin{table*}[t]
\begin{NiceTabular}{l *{5}{m{2cm}}}
\toprule
 \textbf{perturbation setting} & \textbf{AUC-all} & \textbf{AUC-single} & \textbf{AUC-multi} & 
 \textbf{eff diff (single)} &
 \textbf{eff diff (multi)}  \\
\midrule
\makecell[l]{
normal \\
\quad (+ tall, - taller)
}& 
0.889 & 0.981 & 0.796 & 
\makecell{
$3.24 \pm 0.08$ \\ $(p < 0.001)$ 
} &
\makecell{
$1.3 \pm 0.3$ \\ $(p < 0.001)$  \\
} \\

\makecell[l]{
no contrastive \\
\quad (+ tall)
}& 
0.831 & 0.963 & 0.755 & 
\makecell[l]{
$1.96 \pm 0.08$ \\ $(p < 0.001)$ 
} &
\makecell{
$1.0 \pm 0.3$ \\ $(p < 0.001)$  \\
} \\
\midrule

\makecell[l]{no context \\ \quad (+ tall, - taller) } & 
0.792 & 0.995 & 0.681 & 
\makecell{
$12.1 \pm 0.4$ \\ $(p < 0.001)$ 
} &
\makecell{
$2.9 \pm 1.4$ \\ $(p=0.045)$  \\
} \\

\makecell[l]{
common target (blue)\\
\quad (+ blue, - taller)
}& 
0.767 & 0.836 & 0.738 & 
\makecell{
$2.6 \pm 0.1$ \\ $(p < 0.001)$ 
} &
\makecell{
$1.6 \pm 0.5$ \\ $(p < 0.001)$  \\
} \\

\makecell[l]{common target (blue) \\ \quad no contrastive \\ \quad (+ blue)
} & 
0.743 & 0.782 & 0.717 & 
\makecell{
$1.9 \pm 0.1$ \\ $(p < 0.001)$ 
} &
\makecell{
$1.7 \pm 0.5$ \\ $(p < 0.001)$  \\
} \\

\makecell[l]{-eze morpheme \\ \quad (+talleze, - taller)} & 
0.709 & 0.794 & 0.665 & 
\makecell{
$1.2 \pm 0.7$ \\ $(p < 0.001)$ 
} &
\makecell{
$0.5 \pm 0.2$ \\ $(p=0.022)$  \\
} \\

\bottomrule
\end{NiceTabular}
\caption{
Results under various changes to the train and eval remappings on RoBERTa-large.
Row 1 reproduces the result presented in the main text, for comparison.
AUC-single and AUC-multi give clusterability of the subset of words that are single or multi-token.
Eff diff give the difference in effect in the linear model across same-class and between-class, then conditioned by whether words are single or multitoken.
}
\label{tab:app:morpho_additional}
\end{table*}

\paragraph{Mediators of transfer (i.e. ablations)}
As perturbation enables us to easily alter aspects of the context or critical region remappings, we further investigate the extent to which generalization is mediated by different aspects of the remappings (\autoref{tab:app:morpho_additional}).

As we saw that tokenization (single vs. multiple) has a significant effect on transfer, we report results (AUC and average marginal effect) by tokenization.
(For example, AUC-single is the clusterability metric but only on the portion of the similarity matrix corresponding to single-token words.)
We again compute a linear model in the same way as before, but with factors only for \emph{class} and \emph{tokenization}.
Eff-diff in the last two columns gives the \emph{difference} between marginal effects for in-class vs. between-class transfer, separated by single token (second-to-last column) and multi-token (last column) conditions.
Eff-diff reflects whether or not the two classes (deverbal vs. comparative) are distinguished by the model using perturbation.

Row 1 reproduces the main text result for RoBERTa-large.
Row 2 ablates the contrastive loss: normally we train (+tall, -taller), but here we train only to (+tall). 
Evaluation in the no-contrastive setting is still computed with log odds of both items (tall, taller).

In the following rows we explore three remapping changes which are potentially linguistically meaningful. 
\textit{First}, we set the context to $\emptyset$ and perturb only on the critical region.
Single-token words are still well-distinguished (AUC 0.995), but multitoken words have clusterability of only 0.681.
In other words, without context, transfer in single-token pairs still reflects differences in morphological class. 
On the other hand, multi-token words are no longer separated, suggesting that the model may rely more on context to achieve stable representations for multitoken words.
(Note that effects in the no-context setting are larger since perturbation is no longer conditioned on any context.)

\textit{Second}, instead of contrastively perturbing e.g., toward \textit{tall} and away from \textit{taller}, we perturb only towards a different static token, \textit{blue}.
This allows us to assess how the model transfers the perturbation from the context alone, without any direct feedback about the intended content of the critical region.
For completeness, we present this result both with and without contrastive loss; the linguistically meaningful result that captures transfer due to context only is the one without contrastive loss.
This approach gives overall AUC of 0.74, showing that the contexts contain meaningful information that affects how the model generalizes the training remapping.

\textit{Third}, we test transfer using an unknown morpheme (e.g. taller -> talleze).
(Note, in this setting, the evaluation remapping use talleze vs. taller rather than tall vs. taller.)
This produces the lowest AUC.
In other words, perturbing to a novel suffix inhibits transfer along morphological class dimension, possibly because the model is unlikely to have a pre-existing representation for \textit{X-eze}.
(Moreover, the fact that X-eze is always multi-token likely further confounds transfer.)

\subsection{Results for other models}
\label{sec:app:morph:other_models}

\begin{figure*} 
    \centering
    \includegraphics[width=0.8\textwidth]{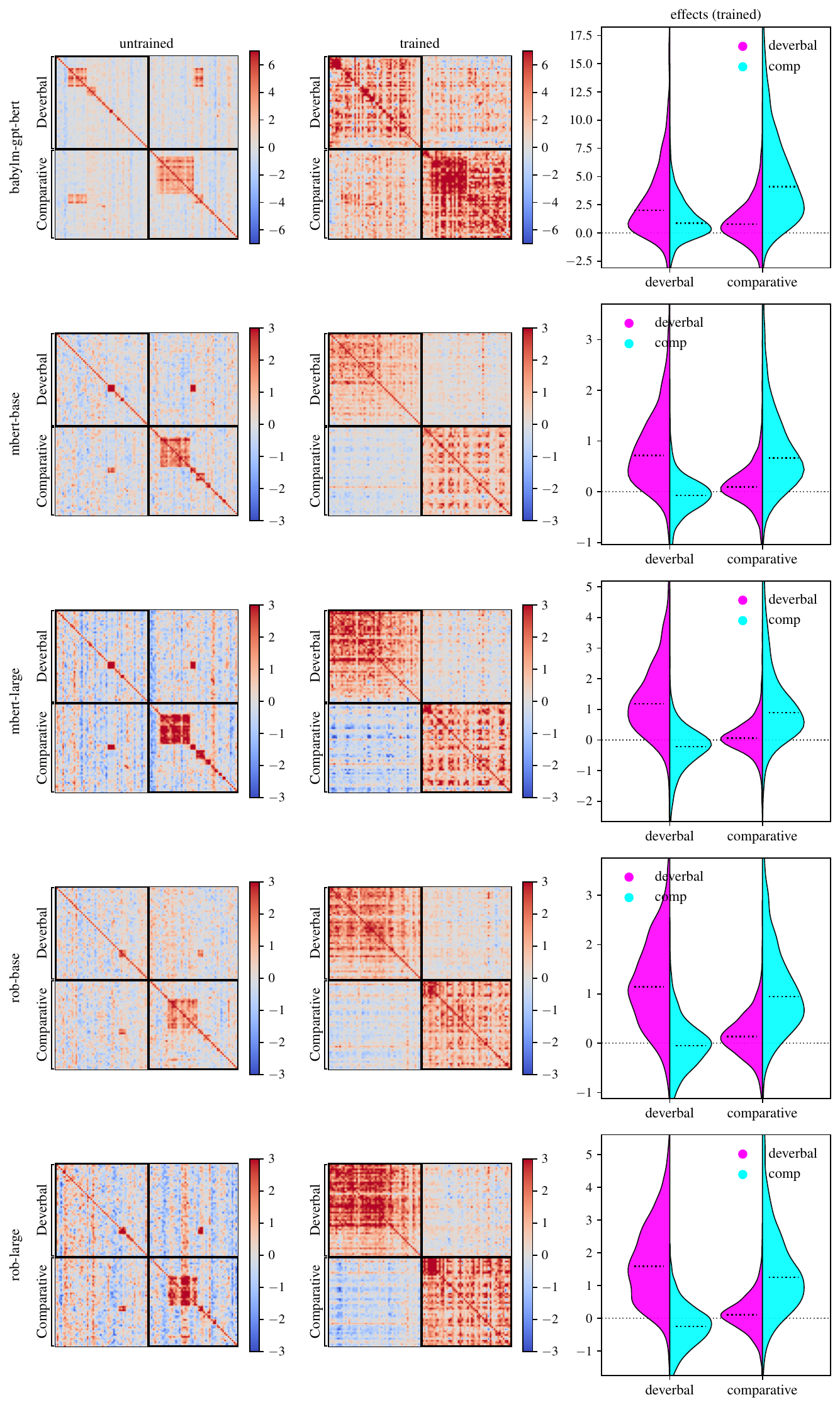} 
    \caption{
    Results for all models, with untrained on left, trained in middle, and effect distribution (log-odds) in trained models on right.
    Effect sizes in the heatmaps are clipped to 7,3,3,3,3 respectively (effect sizes vary based on learning rate, which was tuned per model) so that high transfer values on the diagonal do not obscure the rest of the heatmap structure.
    The final row (rob-large) is the same as shown in the main text (though with different sorting of rows and columns).
    }
    \label{fig:app:morpho_all_models}
\end{figure*}

We use the following parameters for perturbation (chosen by highest clusterability AUC on the train set):
RoBERTa-base: LR=1e-5, steps=5;
RoBERTa-large: LR=1e-5, steps=5;
modernBERT-base: LR=5e-5, steps=3;
modernBERT-large: LR=5e-5, steps=3;
GPT-BERT: LR=5e-4, steps=5.

AUC clusterability across models was presented in the main text (\autoref{tab:morpho}), and transfer-based similarity matrices  are shown in \autoref{fig:app:morpho_all_models}.
We observe that GPT-BERT exhibits relatively smaller average effect size for deverbal-deverbal (row 1, col 3), which may agree with prior work in humans and models that the derivational morphologies like the deverbal (verb to noun) may be harder to generalize than inflectional ones \cite{gladkova-etal-2016-analogy, pacton2023morphology}.

\subsection{Additional details for cosine baseline}
\label{sec:app:morph:cos}

\begin{table*}[t]
\centering
\begin{NiceTabular}{m{2.5cm} *{9}{m{0.65cm}}}
\toprule
\RowStyle{\bfseries}
\Block{2-1}{Model} & \multicolumn{3}{c}{perturb} & \multicolumn{3}{c}{cosine best} & \multicolumn{3}{c}{cosine last} \\
\cmidrule(lr){2-4}
\cmidrule(lr){5-7}
\cmidrule(lr){8-10}
 & all & single & multi &  all & single & multi & all & single & multi \\ 
\midrule
 ModernBERT-base & 0.837 & 0.940 & 0.788  & 0.912 & 0.971 & 0.931 & 0.772 & 0.892 & 0.782 \\
 ModernBERT-large & 0.857 & 0.956 & 0.807 & 0.883 & 0.920 & 0.930 & 0.722 & 0.886 & 0.766 \\
 RoBERTa-base & 0.867 & 0.957 & 0.774 & 0.868 & 0.915 & 0.880 & 0.764 & 0.849 & 0.759 \\
 RoBERTa-large & 0.889 & 0.981 & 0.796 & 0.893 & 0.944 & 0.917 & 0.657 & 0.676 & 0.677 \\
 GPT-BERT & 0.737 & 0.987 & 0.723 & 0.908 & 0.964 & 0.914 & 0.727 & 0.843 & 0.757 \\

\bottomrule
\end{NiceTabular}
\caption{
Morphology: AUC clusterability for perturbation and cosine conditioned on tokenization (all, single, multi).
}
\label{tab:app:morpho:all_cosine}
\end{table*}

In \autoref{tab:app:morpho:all_cosine} we provide a breakdown of cosine baseline results for all models, conditioned on tokenization.
The \emph{all} columns reproduce the values from \autoref{tab:morpho}.
For cosine, these values come from running each model forward with the target word unmasked. 
For example, in ``The man was \textbf{taller} than the boy'' we would extract the contextualized representation for \textbf{taller}.
(In the case of multiple tokens, we take the average.)
This differs from perturbation, where the target word is masked for both perturbation (training) and evaluation.

\autoref{tab:morpho} reflects that \emph{cosine-best} achieves higher \emph{AUC-all} than perturbation. 
Here we see that cosine's outperformance is a result of performance on multitoken words:
Perturbation performs better than \emph{cosine-best} in 4 of 5 models on single-token words.

\section{Supplement for \sectionref{sec:wsd}}
\label{app:wsd}

\subsection{Methods}
\label{app:wsd:methods}

\paragraph{Dataset}
We provide additional details for the CoarseWSD-20 dataset (CWSD; \citealt{loureiro-etal-2021-analysis}).
All sentences in the dataset are entirely lowercase.
The dataset consists of the following words: 
apple, arm, bank, bass, bow, chair, club, crane, deck, digit, hood, java, mole, pitcher, pound, seal, spring, square, trunk, yard.

When sampling sentences for our train and test sets, we omit a small number of examples where the target word occurs multiple times in the sentence.
As noted in \sref{sec:wsd}, for each of the 20 words in the dataset, we sample 100 examples equally balanced over the senses for that word.
Some words have insufficient examples for a given sense.
In those cases, clusterability is reweighted to be balanced (see \aref{app:methods:auc}).
The total number of examples sampled is 1606 (versus 2000 possible), with 136k similarity scores for analysis.

As we find that tokenization can impact perturbation results (\sref{sec:morpho}), we report words that are not single-token:
For RoBERTa, 5 usages in the test dataset have the target word split into two tokens (this can happen, for example, if the target word occurs at the start of the sentence). 
For modern-BERT, in addition to those 5 usages, all usages of `crane' are two tokens.
For GPT-BERT, `crane', `hood', `java', `mole', and `pitcher' are always two tokens.

\paragraph{Hyperparameters}
Hyperparameters (learning rate and number of optimizer steps) were tuned on separate data for training (the CWSD train split) to maximize average AUC clusterability across word senses.
For the final test, we use the following settings:
RoBERTa-base: LR=3e-6, steps=3;
RoBERTa-large: LR=3e-6, steps=1;
modernBERT-base: LR=3e-5, steps=3;
modernBERT-large: LR=3e-5, steps=3;
GPT-BERT: LR=1e-6, steps=1.

\paragraph{Computations}
For the word-vector baseline, we run the model forward and extract the hidden state.
When a word has multiple tokens, we follow \citet{teglia-etal-2025-much} and average.
We then compute the similarity matrix using cosine similarity.

\subsection{Additional results: other perturbation targets}
\label{app:wsd:nonce}
We test three other remapping targets on modernBERT-large for the test set (using the same hyperparameters obtained from tuning against ``glam''); see \autoref{app:tab:nonce_words}.

\begin{table}[t]
\centering
\begin{NiceTabular}{lr}
\toprule
\RowStyle{\bfseries}
Word & Perturb AUC \\
\midrule
glam & 0.871 \\
aaaa & 0.883 \\
acuity & 0.865 \\
nonce & \textbf{0.889} \\
\bottomrule
\end{NiceTabular}
\caption{
Average AUC on test set using other perturbation targets in modernBERT-large.
}
\label{app:tab:nonce_words}
\end{table}

\subsection{Similarity visualizations}
\label{app:wsd:wsd_viz}

See \autoref{fig:app:wsd_sims}.
Perturbation provides both quantitatively higher AUC and qualitatively better visual separation between word senses.
Perturbation also more clearly reveals graded relatedness of different senses.
For example, for \textit{square}, there is visibly more transfer between the geometric shape usage and the town square usage (which share a notion of shape) than there is between either of these usages and the company usage (which has little semantically in common with the geometric notion of a square).

\section{Supplement for \sectionref{sec:fg}}
\label{app:fg}

\subsection{Methods}

\begin{table*}[t]
\footnotesize
\centering
\setlength{\tabcolsep}{2pt}
\renewcommand{\arraystretch}{1.1}
\begin{tabular}{r|lllllll}
\toprule
\textbf{Construction} & \textbf{Prefix} & \textbf{Filler} & \textbf{NC} & \textbf{Article} & \textbf{NP} & \textbf{Verb} & \textbf{Label}
\\
\midrule
Emb.~Wh-Q (\textit{Know})   & I know   & who/that &    & the & man & liked & ./him   \\
\mbox{Emb. Wh- (\textit{Wonder})} & I wonder & who/if   &    & the & man & liked & ./him   \\
Matrix Wh-Q                        &         & Who/""   & did& the & man & like  & ?/him   \\
Restr. Rel. Clause               & The boy & who/and  &    & the & man & liked & was/him \\
Cleft                                     & It was  & the boy/clear & that & the & man & liked & ./the boy \\
Pseudo-Cleft                              &         & Who/That &    & the & man & liked & was/it \\
Topicalization                            & Actually, & the boy/"" &  & the & man & liked & ./the boy \\
\midrule
Agreement Control (\ctla) & The     & boy/boys & that & the & man & liked & is/are \\
Transitive Control (\ctlt)               &         & Once/Today & & some/that & man/boy & ran/liked & ./him \\
\bottomrule
\end{tabular}
\caption{Studied filler-gap pairs, reproduced from \citet{boguraev-etal-2025-causal}}
\label{tab:app:bog_tab}
\end{table*}

\paragraph{Dataset}
\autoref{tab:app:bog_tab} is reproduced from \citet{boguraev-etal-2025-causal} and shows example non-embedded, animate minimal pairs.
For examples of the inanimate and embedded conditions, the reader is referred to the original paper.

\paragraph{Perturbation}
During perturbation training and evaluation, for all completions/targets (sentence suffixes) in the \citeauthor{boguraev-etal-2025-causal} dataset, only the first token of any completion is considered. 
Though some examples in the dataset have multiple word targets, the authors confirmed (p.c.) that only the first (single-token) word is considered in these cases.

The perturbation code released with our work is implemented so that multiple token targets can be targeted.
Though we test multitoken targets in the MLM case, we do not do so in the autoregressive case, to improve comparability to prior work.

\paragraph{Experiment}
For each trial, we sample a single perturbation target and 25 disjoint evaluation examples.
For each of of the 34 conditions (7 FG $\times$ 2 animacy $\times$ 2 embeddedness, plus 2$\times$2 cross of animacy and embeddedness conditions for \ctlt, plus 2 embeddedness conditions for \ctla), we run 5 separate trials (34 $\times$ 5 total perturbations).

Because we are trying to perturb the model's representation of FG constructions, we always perturb to the incorrect completion in a minimal pair (e.g., \emph{I know who the man liked \textbf{him}}, \emph{The boy that the man liked \textbf{are}}, \emph{Once some man ran \textbf{him}}).
For these \emph{training} perturbations, 
the prefix string is always
\begin{itemize}
    \item  for FG (vs non-FG): FG (e.g., who)
    \item for \ctla: singular (e.g., boy)
    \item for \ctlt: intransitive (e.g., ran).
\end{itemize}
(These all parallel the structure of \autoref{tab:app:bog_tab}.)

For evaluation, we use both versions of the prefix string for FG (these form our FG, nonFG conditions).
For evaluation we do not use the \emph{prefix} strings for plural \ctla nor transitive \ctlt.

\paragraph{Hyperparameters}
For hyperparameter tuning (learning rate and number of optimizer steps), we maximize AUC clusterability, where we take as our classes FG and controls (\ctla, \ctlt).
We do not consider whether the model groups different controls together, just whether the entire group \ctla and \ctlt is more self-similar than similar to FG.
Final hyperparameters for Pythia-1.4B are LR=1e-6 and steps=5.

\subsection{Effects of Critical Region Similarity}
\label{app:fg:jaccard_sim}
We compute \crs (\crshort) for the minimal pairs using a modified Jaccard measure.
The similarity of any two conditions $S(i,j)$ is given by
\[
S(i, j)
=
\frac{
\displaystyle \sum_{w \in \mathcal{V}}
\min\!\big(C_i(w),\, C_{j}(w)\big)
}{
\max\!\left(
T_i, T_j 
\right)
}
\]
where $\mathcal{V}$ is the space of all possible label words,
$C_i(w)$ is the count of the word $w$ in the minimal pairs for condition $i$, 
and $T_i = \sum_{w\in \mathcal{V}} C_i$ is the total size of the label space for class $i$.

We compute the modified Jaccard similarity on the train split (the test split has similar \crshort) for the FG targets/completions (\autoref{fig:app:fg_label_sim_fg}), the non-FG targets/completions (\autoref{fig:app:fg_label_sim_nonfg}), and both together (\autoref{fig:app:fg_label_sim_both}).
In \citet{boguraev-etal-2025-causal}, during training, DAS learns parameters such that the model prefers the FG label (\autoref{fig:app:fg_label_sim_fg}).
During evaluation, the effect of the intervention learned by DAS is evaluated simultaneously against both the FG and non-FG targets.

\paragraph{Observations and implications}
As noted in the main text, in the FG condition, \ctla has no output targets in common with pseudocleft or restricted relative clause (i.e., \crshort is 0), and is thus an anticonservative control. 
In their Section 5, \citeauthor{boguraev-etal-2025-causal} study the degree of transfer between different FG conditions, but they do not control for this confound.

We note that while our \emph{baselined-transfer} condition should account for differences in \crshort, we may nonetheless still observe patterning in filler-gap similarities that resemble the \crshort matrix. 
This is a strength of our design: we want to be able to capture deeper representational similarity even between construction types that are similar for more superficial reasons, and our controls enable us to do this.

\begin{figure}[t]
    \centering
    \includegraphics[width=\linewidth]{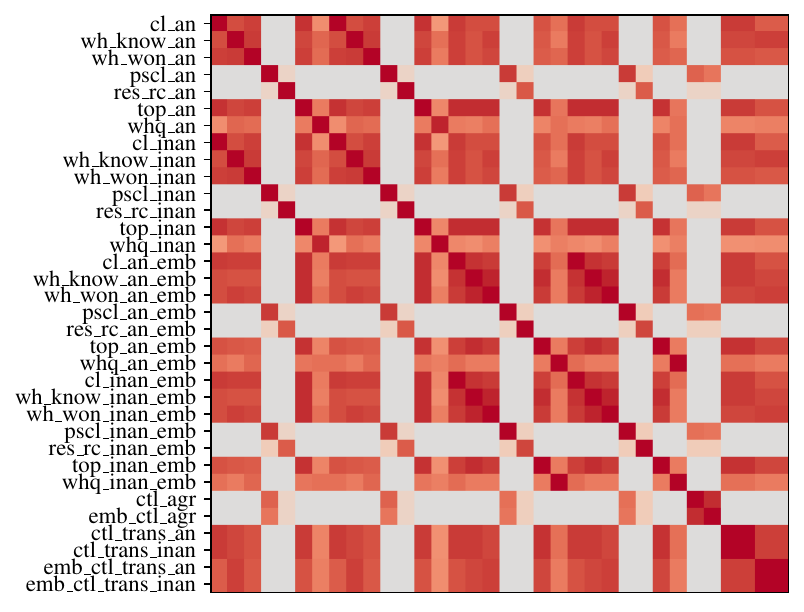}
    \caption{Modified Jaccard similarity of output labels (\crshort) for minimal pairs across conditions for \emph{FG} condition.
    This corresponds to the target labels toward which perturbation and DAS train, and to \emph{positive} log-odds evaluation effects.
    }
    \label{fig:app:fg_label_sim_fg}
\end{figure}
\begin{figure}[t]
    \centering
    \includegraphics[width=\linewidth]{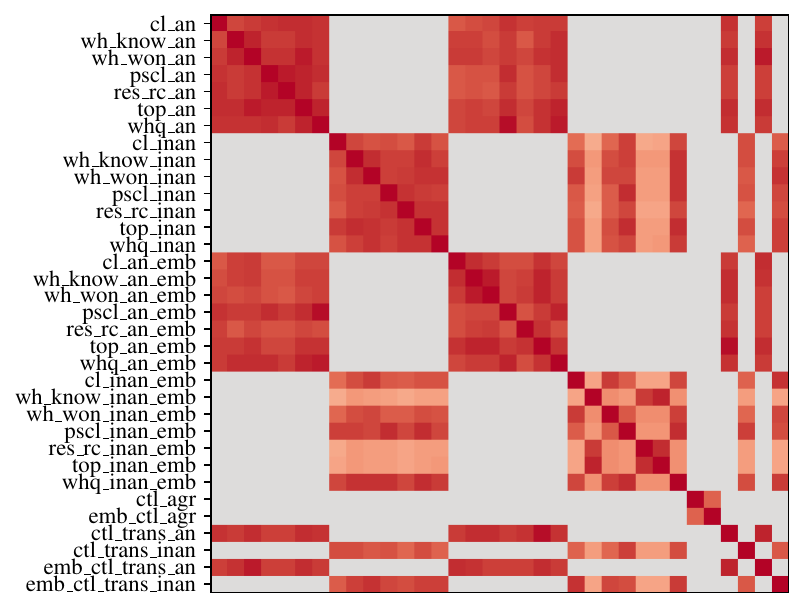}
    \caption{Modified Jaccard similarity of output labels (\crshort) for minimal pairs across conditions for the \emph{non-FG} condition (and \ctla plural, \ctlt transitive).
    This reflects similarity of the contrastive / anti-loss labels for perturbation. 
    This affects \emph{negative} log-odds evaluation effects.
    }
    \label{fig:app:fg_label_sim_nonfg}
\end{figure}
\begin{figure}[t]
    \centering
    \includegraphics[width=\linewidth]{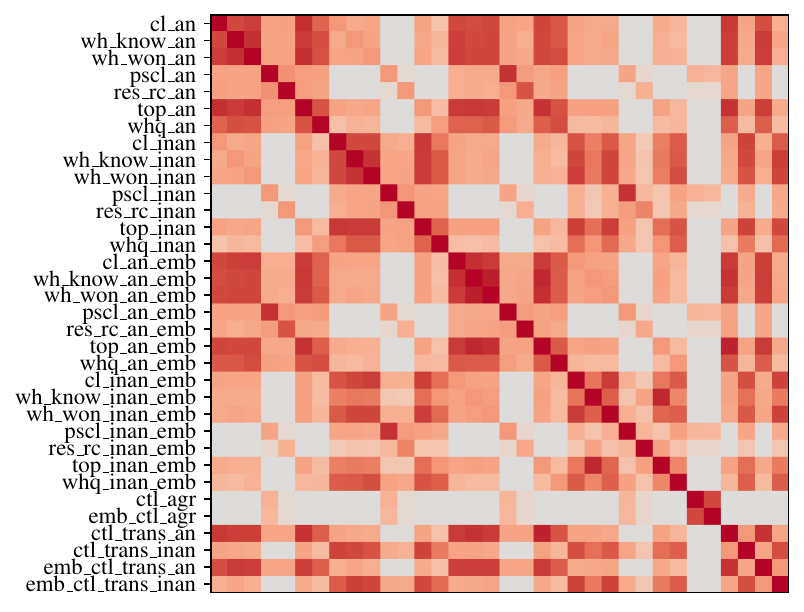}
    \caption{Modified Jaccard similarity of output labels (\crshort) for minimal pairs across conditions aggregating both FG and non-FG labels.
    }
    \label{fig:app:fg_label_sim_both}
\end{figure}

\subsection{Additional discussion of transfer over the course of training}
\label{app:fg:disc}
In \autoref{fig:fg_aggs}, FG and non-FG are seen to be nearly perfectly confounded through step 512.
As the prefix strings differ only in, e.g., \emph{who} vs. \emph{that}, transfer from FG to non-FG is likely mediated by token set overlap before the FG/non-FG distinction is learned.

Through step 512, we also see transfer to/from \ctlt.
Note that this corresponds to transfer to/from the perturbation \perturb{I know who the man liked}{.}{}{him} from/to \perturb{Once some man ran}{.}{}{him}.
In this case, excess transfer between FG and \ctlt could be mediated by confounding of transitive and intransitive verbs or by the absence of an FG-nonFG distinction. 
We did not investigate the extent to which the model confounds transitive and intransitive verbs early on. 
We note that even without an FG-nonFG distinction, as the model learns to distinguish transitive and intransitive verbs, we would expect transfer from FG conditions (which all use transitive verbs) to \ctlt to decrease.

\subsection{Statistical comparisons}
\label{app:fg:stat}
\begin{table}[h]
\begin{NiceTabular}{p{1cm} m{0.8cm} m{1.2cm} c c}
\toprule
\textbf{condition} & \textbf{trained} & \textbf{class} &
\textbf{effect} & \textbf{p-val} \\
\midrule

\Block{6-1}{animacy}
  & \Block{3-1}{false}
      & diff   & 0.001& 0.73 \\
  &   & same\_an   & 0.014 & 0.002 \\
  &   & same\_inan & 0.003 & 0.49 \\
\cmidrule(lr){2-5}
  & \Block{3-1}{true}
      & diff   & 0.773 & $<0.001$  \\
  &   & same\_an   & 1.358 & $<0.001$ \\
  &   & same\_inan & 1.037 & $<0.001$ \\
\midrule
\midrule
\Block{6-1}{embed}
  & \Block{3-1}{false}
    & diff   & 0.006 & 0.07 \\
  &   & same\_emb   & -0.004 & 0.43 \\
  &   & same\_unemb & 0.016 & $<0.001$ \\
\cmidrule(lr){2-5}
  & \Block{3-1}{true}
    & diff   & 0.687 &$<0.001$   \\
  &   & same\_emb   & 1.356 & $<0.001$\\
  &   & same\_unemb & 1.124 & $<0.001$\\

\bottomrule
\end{NiceTabular}
\caption{
Syntax (filler-gap) Linear Model: Average effects for \blt by model trained status and animacy and embeddedness condition.
}
\label{tab:app:fg_lm}
\end{table}

We present additional statistical analyses.

\paragraph{Transfer to/from FG is significant}
First, we test the significance of observed differences in \figref{fig:fg_aggs}{B}.

We tested whether mean FG->FG effect is greater than FG->non-FG and FG->control effects using a one-sided permutation tests (n = 5,000 permutations). 
FG->FG effects are significantly larger than both FG->non-FG ($\Delta= 0.90, p < 0.001$) and FG->control ($\Delta = 1.18, p < 0.001$).

We also tested whether the effects are significantly greater in trained than untrained models (i.e. for each pair of violin plots):
All were significant ($p<0.001$) with average difference (trained - untrained):
FG->FG $\Delta=1.55$,
FG->non-FG $\Delta=0.65$,
FG -> ctl $\Delta=0.21$,
ctl->FG $\Delta = 0.36$,
ctl -> non-FG $\Delta = 0.20$

\paragraph{Effects of animacy and embeddedness}
In the original \citeauthor{boguraev-etal-2025-causal} study as well as our own, the animacy and embeddedness manipulations were primarily intended as controls to test whether abstract FG transfer occurred across these manipulations.
However, \citeauthor{boguraev-etal-2025-causal} report that these control manipulations have a substantial impact on transfer in and of themselves (i.e., DAS transfer is higher when the items match vs. mismatch the animacy of the extracted element).
Here we explore whether perturbations in this dataset also transfer along animacy and embeddedness dimensions.
Within the FG conditions, we test aggregate \blt across animacy and embeddedness conditions quantitatively using linear regression with three factors: 
\textit{animacy} has three levels: same-animate, same-inanimate, and different;
\textit{embeddedness} has three levels: same-emb, same-unemb, and different;
\textit{trained} is true or false.
We again perform post-estimation using the \margeff package.
The response variable is the unaggregated cell values (log-odds ratio) from the matrices in \autoref{fig:app:fg_acq_all}, using the third column, which is \blt, and the first row (step 0, untrained) and the last row (step 143k, trained), for a total of 196k observations.

Marginal effects are given in \autoref{tab:app:fg_lm}.
In all but one case, effects are not significantly different from 0 for untrained models. 
All effects in trained models are significant ($p< 0.001$).

Within trained models, the effect of same vs. different animacy and embeddedness is significant: 
the marginal effect of same animacy vs. diff animacy is 0.42, $p<0.001$ and same embeddnedness vs. diff embeddedness is 0.55, $p<0.001$.

\subsection{Comments on FG similarities}
\label{app:fg:fg_sims}

\begin{figure}
    \centering
    \includegraphics[width=\linewidth]{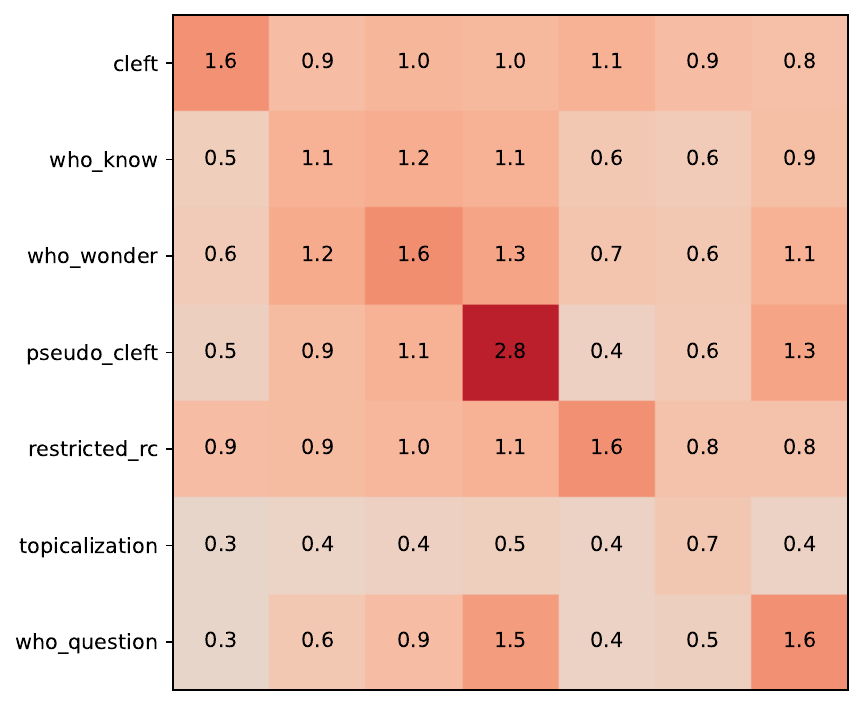}
    \caption{
    Average \blt in Pythia-1.4B aggregated by FG condition.
    This corresponds to an aggregate (averaged) of the final step 143k FG-nonFG plot in \autoref{fig:app:fg_acq_all}.
    }
    \label{app:fg:fig:aggregates}
\end{figure}

While we have argued for a common core shared representation across the FG conditions, we have not studied representational differences/similarities between the different FG conditions themselves.
While highly relevant to linguistic theory \cite{sag2010english}, a full treatment of this question is beyond the scope of this work, and here we simply offer some exploratory observations.
In \autoref{app:fg:fig:aggregates} we show aggregated (across animacy and embeddedness conditions) \blt.
We make the following observations: 
first, topicalization shows the lowest transfer in and out.
Second, self-transfer in pseudoclefts is notably higher than for other conditions.
Third, off-diagonal, the most substantial transfers ($>1.1$) are who\_question to/from pseudocleft (1.5,1.3); 
who\_nonfinite to pseudocleft (1.3);
wh\_finite to/from wh\_nonfinite (1.2, 1.2).
Further work is needed substantiate these observations and link them to theoretical accounts of linguistic representations.

\section{Use of AI Assistants}
ChatGPT and Claude (AIs) were used to produce initial versions of some visualization code.
Some code was initially generated by AI for human review.
All code (including from AIs) used in this study was carefully tested and reviewed by the authors.

\begin{figure*}[h] 
    \centering
    \includegraphics[width=\textwidth]{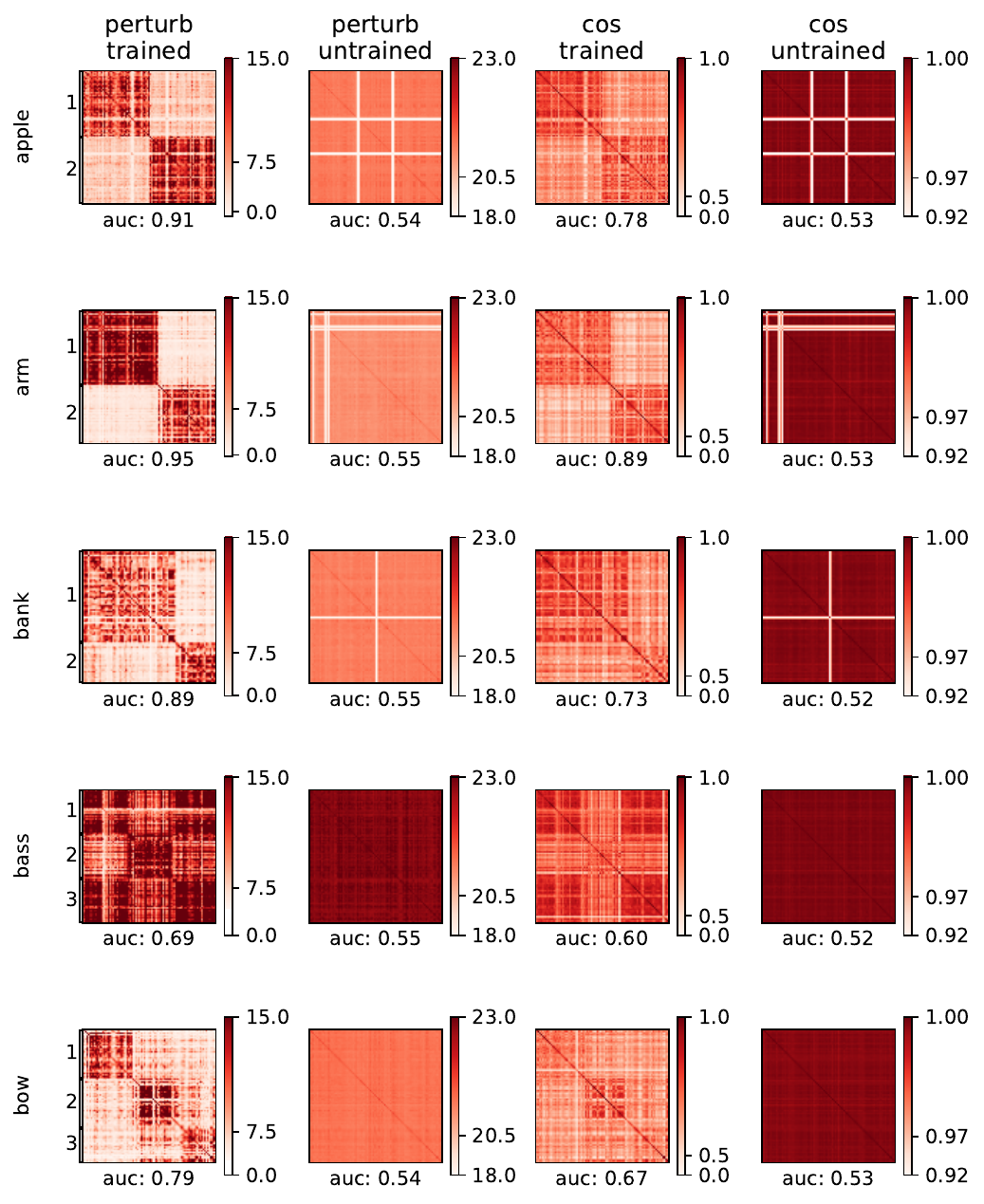} 
    \caption{
    Lexical Representations: Similarity matrices for all words with modernBERT-large using last-layer cosine.
    \textbf{Clipping:} Values for perturbation trained are clipped to 15 so that the diagonal does not overwhelm visualization.
    Values for perturb untrained and cosine untrained are clipped to show as much structure as possible. 
    The \textbf{color scale} is nonlinear to enhance visualize distinction (computed with PowerNorm).
    Cross patterns in untrained models result from differences in tokenization (e.g., when the target word occurs at the beginning of the sentence).
    We note that the three senses for \emph{bass} are bass guitar, bass (vocal), and bass (string instrument).
    \emph{bow} has three senses: bow of a boat, bow (and arrow), bow for string instrument.
    The second and third senses seem to interact; both are bow-shaped.
    }
    \label{fig:app:wsd_sims}
\end{figure*}
\begin{figure*}[h!] 
    \ContinuedFloat
    \centering
    \includegraphics[width=\textwidth]{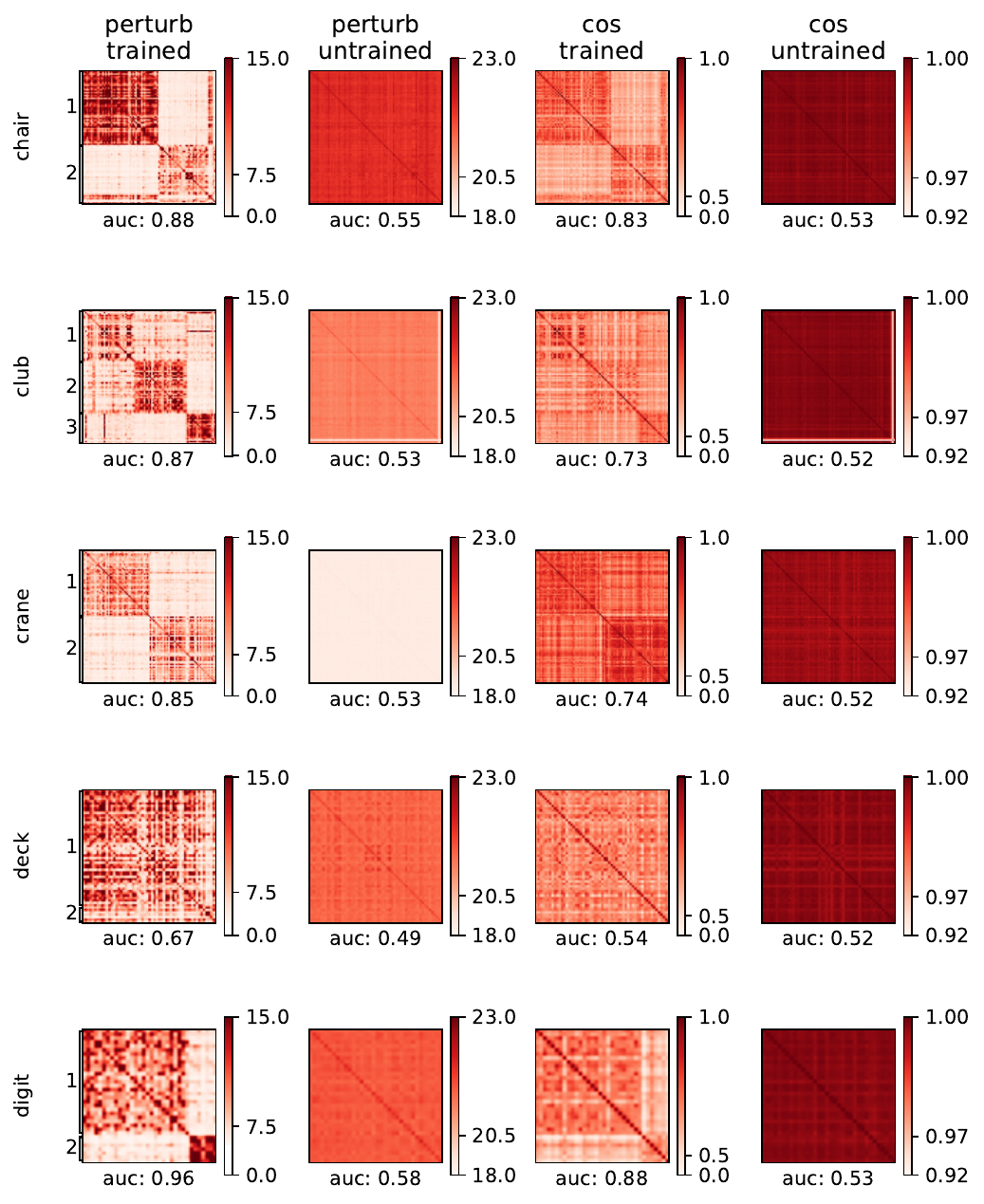} 
    \caption{
    (Fig contd). The apparently anomalous bottom 3 rows/cols for \emph{chair} (seat vs. e.g., academic position) are mislabeled. 
    \emph{deck} has only two senses (deck of boat, deck as of a house), with the second having only 7 examples in the test dataset.
    }
\end{figure*}
\begin{figure*}[h!] 
    \ContinuedFloat
    \centering
    \includegraphics[width=\textwidth]{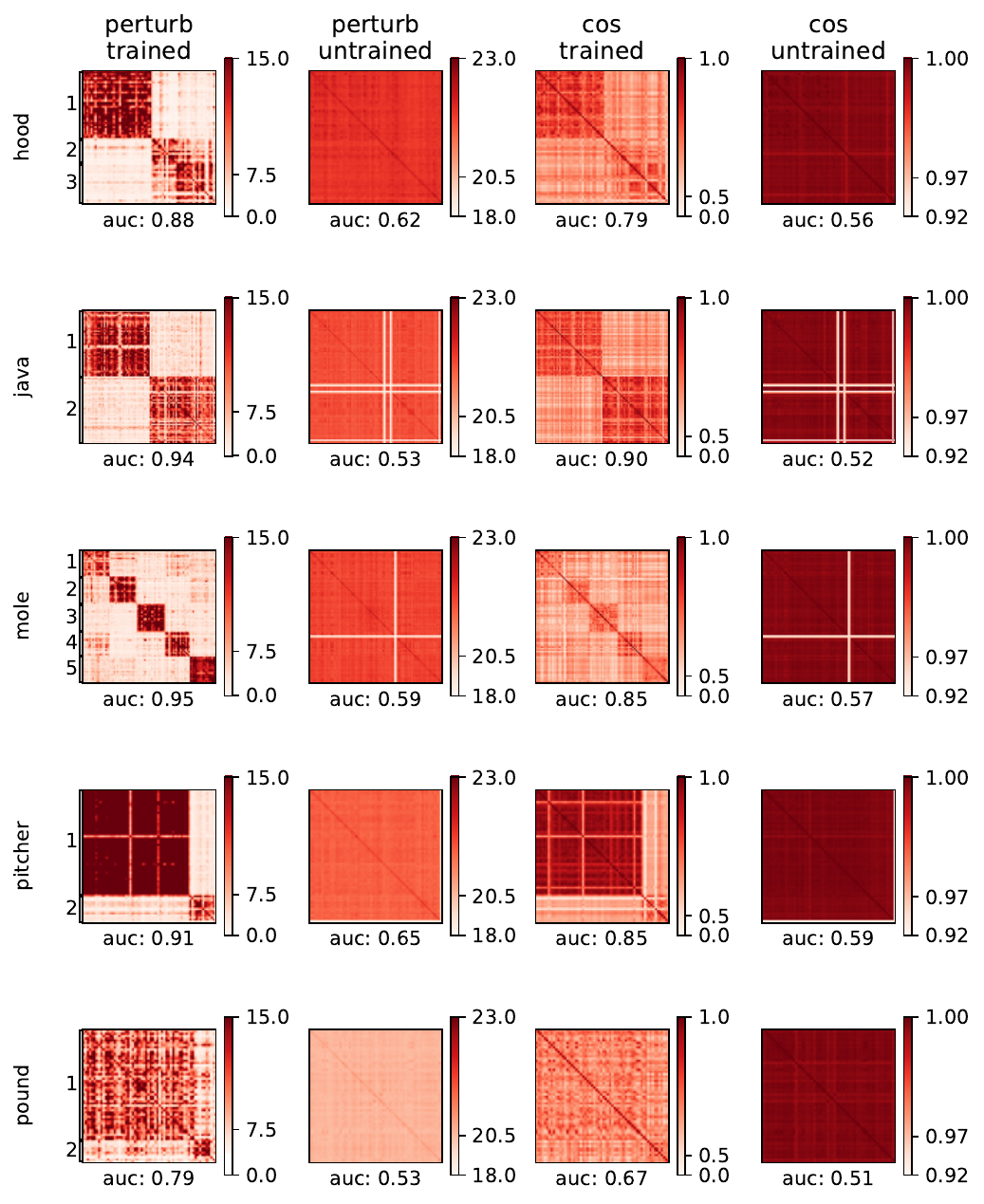} 
    \caption{ (Fig contd).  }
\end{figure*}
\begin{figure*}[h!] 
    \ContinuedFloat
    \centering
    \includegraphics[width=\textwidth]{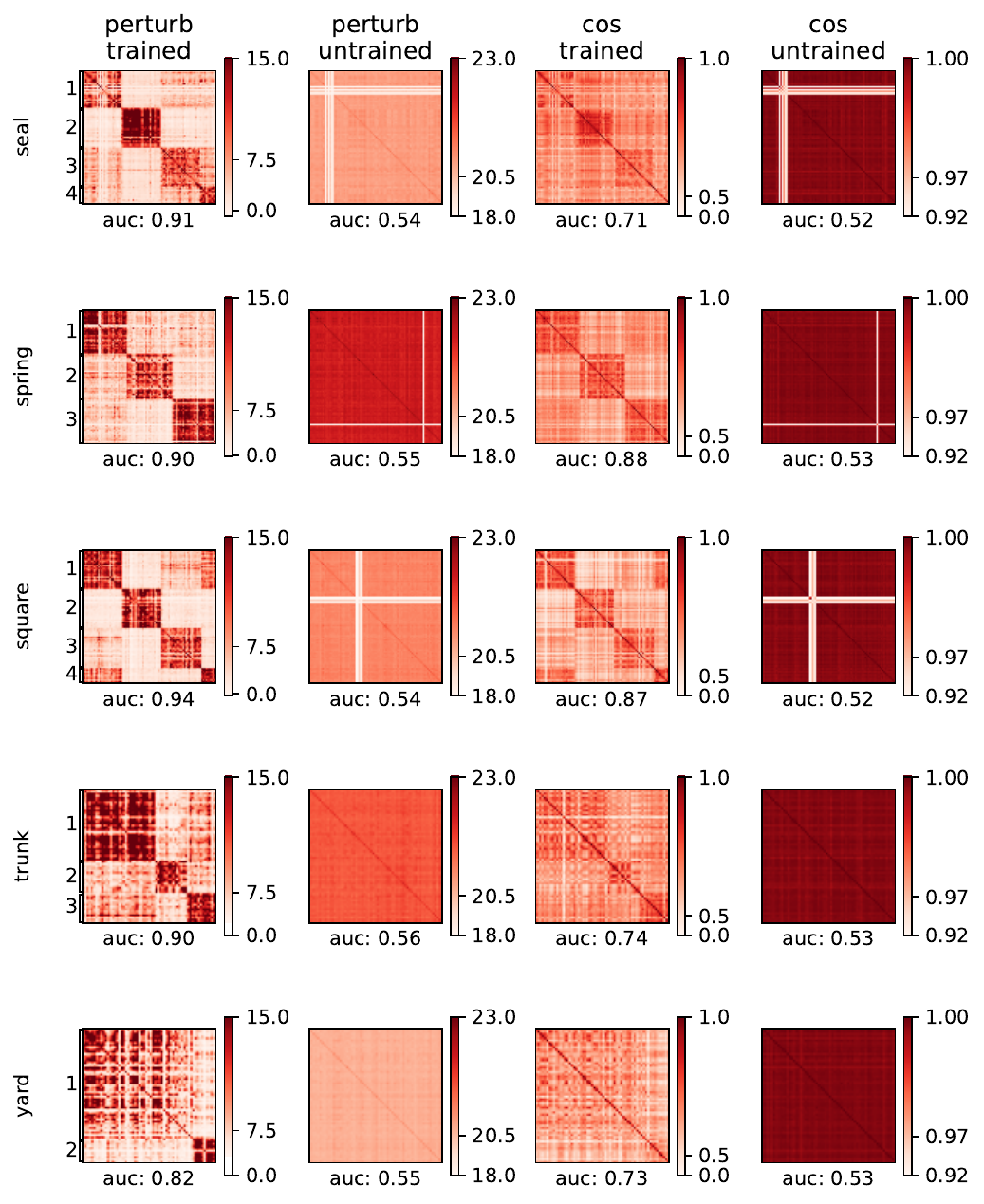} 
    \caption{
    (Fig contd). 
    The four senses of \emph{seal}: animal, R\&B singer, mark, mechanical. 
    The senses for \emph{square} were discussed in the main text.
    The cross pattern is the result of differences in tokenization: examples forming the cross pattern have \emph{square} at sentence start so the token has no included starting space character.
    }
\end{figure*}

\begin{figure*}[h!] 
    \centering
    \includegraphics[width=0.99\textwidth]{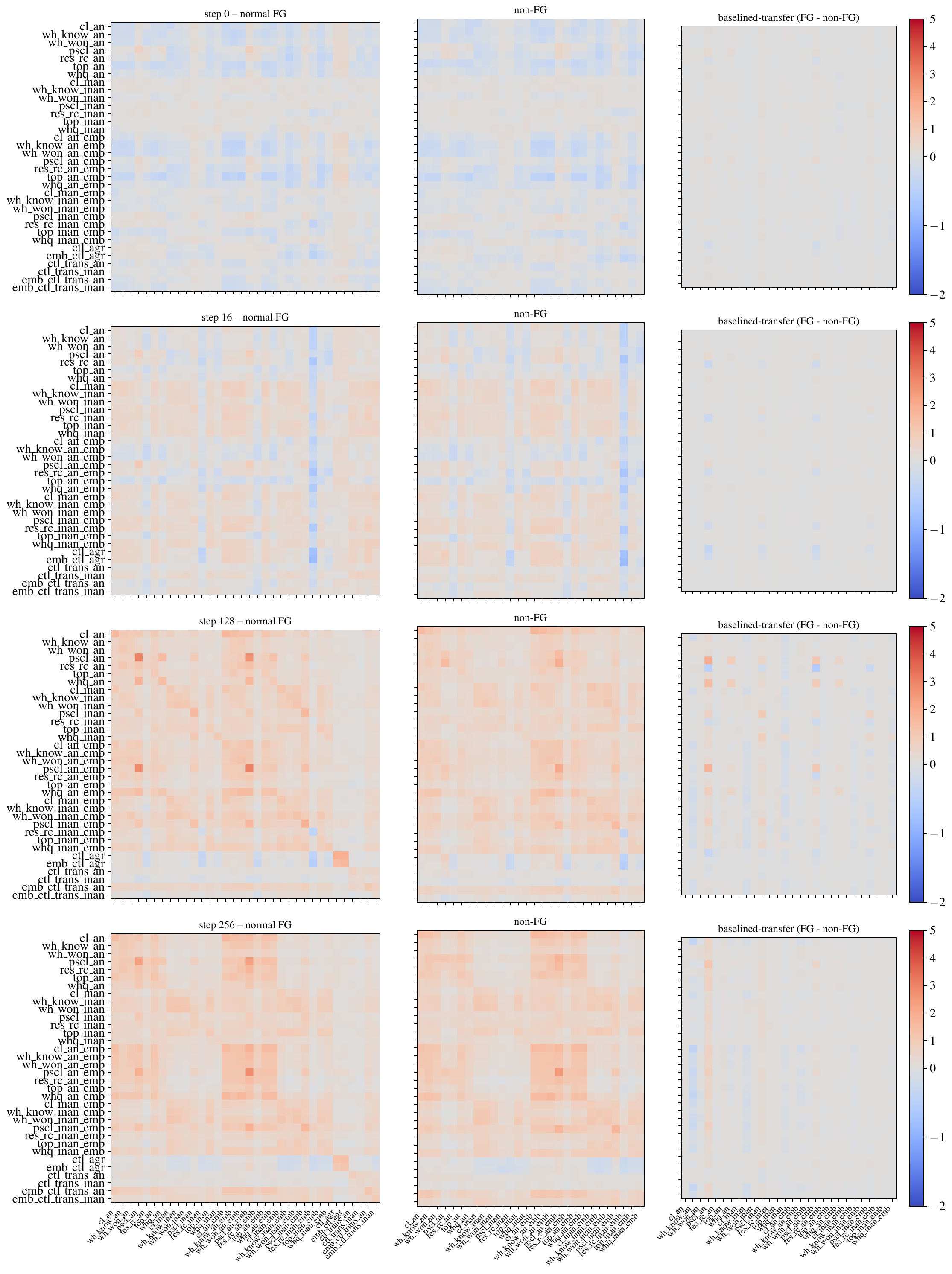} 
    \caption{
    Syntactic Representations:
    Plots over the course of Pythia-1.4b training. 
    Column one (FG) is non-\blt, column two (non-FG) is transfer to the minimal non-FG control, and column three gives \blt.
    As non-FG tests transfer only from FG $\rightarrow$ nonFG, the rightmost 6 columns from the FG matrices are not present (and similarly for \blt).
    Patterning in early steps is clearly seen, which reflects the non-independence of the underlying output label sets.
    Each cell is an average of 125 effect scores from 5 trials each with 25 evaluation remappings.
    }
    \label{fig:app:fg_acq_all}
\end{figure*}
\begin{figure*}[h!] 
    \ContinuedFloat
    \centering
    \includegraphics[width=\textwidth]{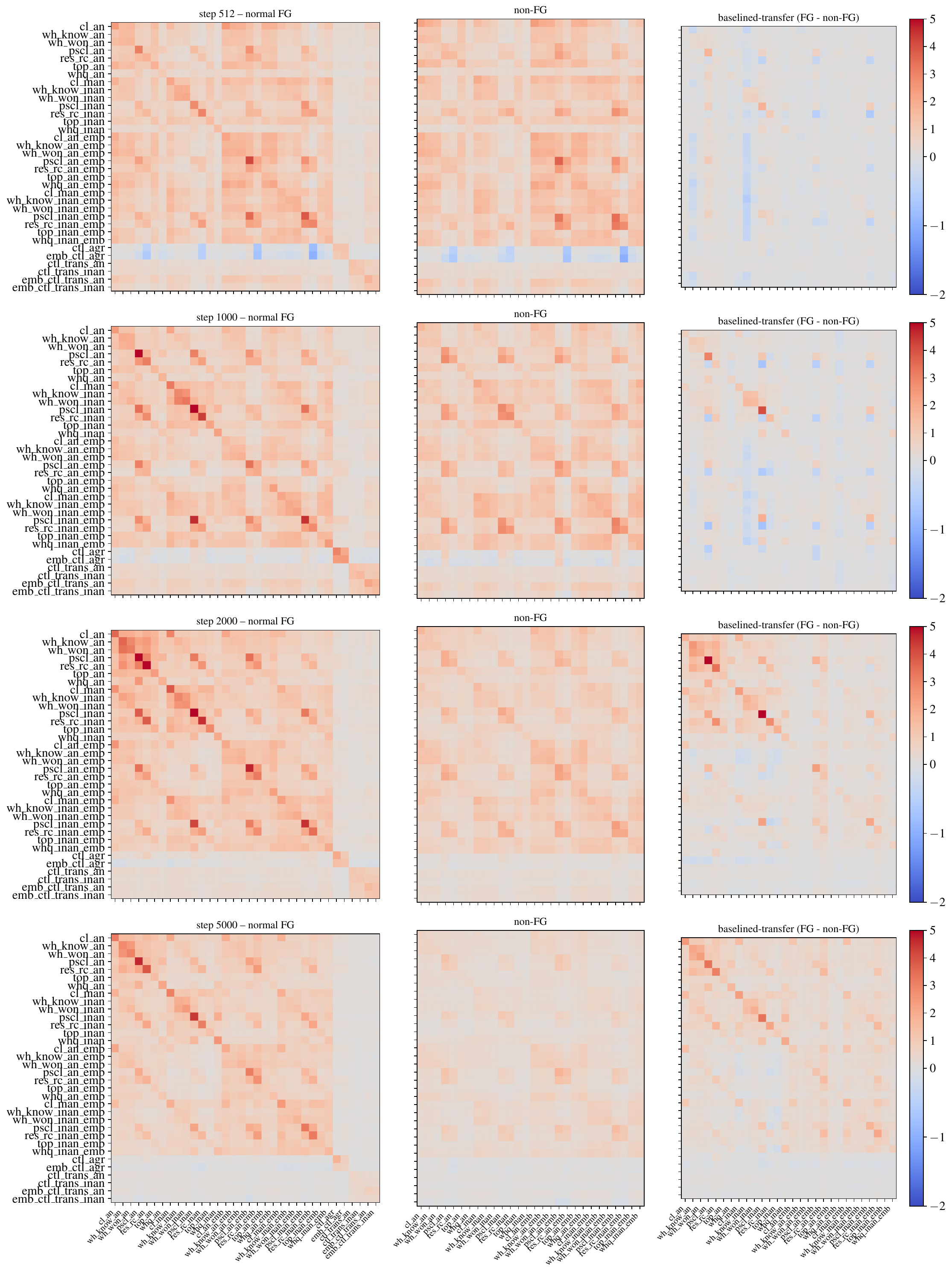} 
    \caption{ (Fig contd).  }
\end{figure*}
\begin{figure*}[h!] 
    \ContinuedFloat
    \centering
    \includegraphics[width=\textwidth]{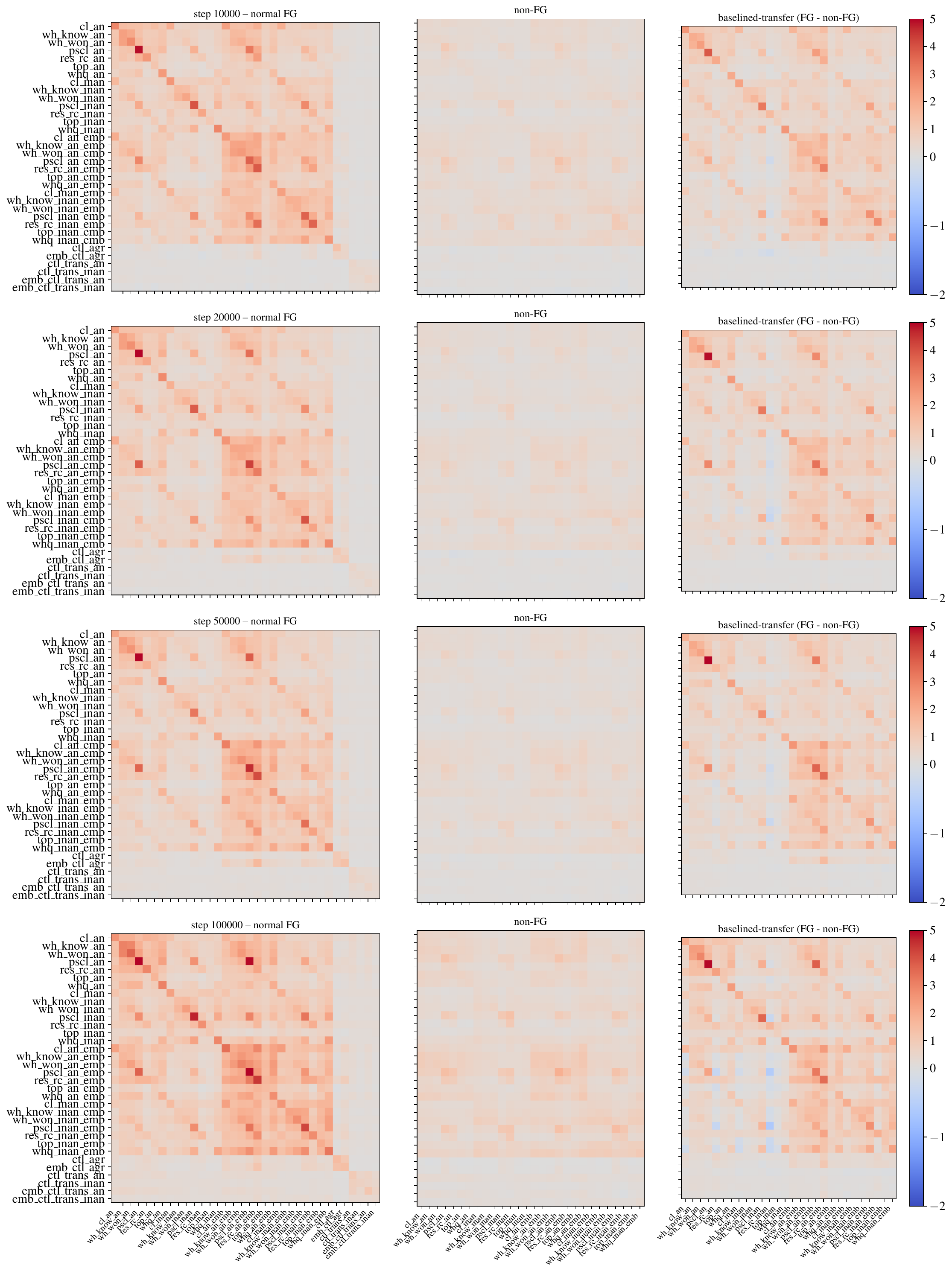} 
    \caption{ (Fig contd).  }
\end{figure*}
\begin{figure*} 
    \ContinuedFloat
    \centering
    \includegraphics[width=\textwidth]{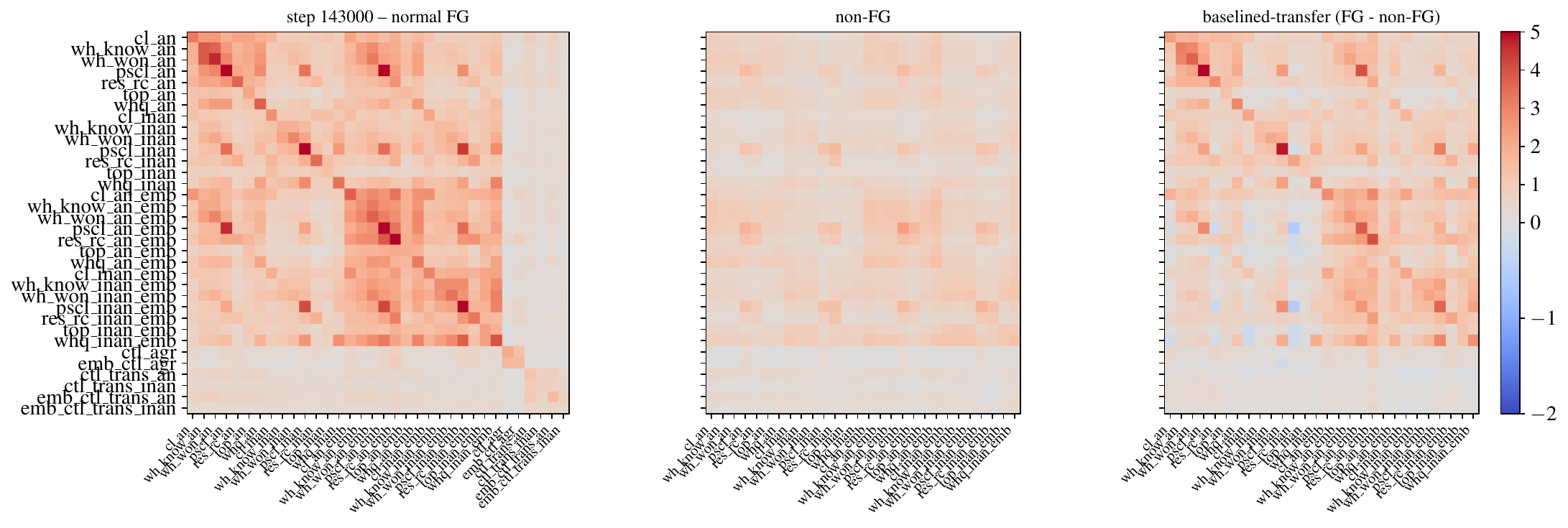} 
    \caption{ (Fig contd).  }
\end{figure*}

\end{document}